\documentclass[final]{cvpr}

\usepackage{times}
\usepackage{epsfig}
\usepackage{graphicx}
\usepackage{amsmath}
\usepackage{amssymb}

\usepackage{booktabs}       
\usepackage{amsfonts}       
\usepackage{nicefrac}       
\usepackage{microtype}      
\usepackage[font=small, skip=8pt]{caption}
\usepackage{amsmath}
\usepackage{amssymb}
\usepackage{multirow}
\usepackage{stackengine} 
\newcommand\oast{\stackMath\mathbin{\stackinset{c}{0ex}{c}{0ex}{\ast}{\bigcirc}}}

\usepackage{wrapfig}
\usepackage{subcaption}
\usepackage{enumitem}
\usepackage{stfloats}

\usepackage[toc,page]{appendix}

\usepackage[pagebackref=true,breaklinks=true,colorlinks,bookmarks=false]{hyperref}

\def\cvprPaperID{7907} 
\def\confYear{CVPR 2021}
\begin{document}

\title{KOALAnet: Blind Super-Resolution using \\Kernel-Oriented Adaptive Local Adjustment}

\author{Soo Ye Kim\footnotemark[1] \qquad\qquad Hyeonjun Sim\thanks{Both authors contributed equally to this work.} \qquad\qquad Munchurl Kim\thanks{Corresponding author.}\\
[0.5em]
Korea Advanced Institute of Science and Technology\\
{\tt\small \{sooyekim, flhy5836, mkimee\}@kaist.ac.kr}
}

\maketitle


\begin{abstract}
  Blind super-resolution (SR) methods aim to generate a high quality high resolution image from a low resolution image containing unknown degradations. However, natural images contain various types and amounts of blur: some may be due to the inherent degradation characteristics of the camera, but some may even be intentional, for aesthetic purposes (e.g. Bokeh effect). In the case of the latter, it becomes highly difficult for SR methods to disentangle the blur to remove, and that to leave as is. In this paper, we propose a novel blind SR framework based on kernel-oriented adaptive local adjustment (KOALA) of SR features, called KOALAnet, which jointly learns spatially-variant degradation and restoration kernels in order to adapt to the spatially-variant blur characteristics in real images. Our KOALAnet outperforms recent blind SR methods for synthesized LR images obtained with randomized degradations, and we further show that the proposed KOALAnet produces the most natural results for artistic photographs with intentional blur, which are not over-sharpened, by effectively handling images mixed with in-focus and out-of-focus areas.
\end{abstract}
\vspace{-0.1em}

\section{Introduction}
When a deep neural network is trained under a specific scenario, its generalization ability tends to be limited to that particular setting, and its performance deteriorates under a different condition. This is a major problem in single image super-resolution (SR), where most neural-network-based methods have focused on the upscaling of low resolution (LR) images to high resolution (HR) images solely under the \textit{bicubic downsampling} setting \cite{kim2016accurate, ledig2017photo, lim2017enhanced, wang2018esrgan}, until very recently. Naturally, their performance tends to severely drop if the input LR image is degraded by even a slightly different downsampling kernel, which is often the case in real images \cite{shocher2018zssr}. Hence, more recent SR methods aim for \textit{blind} SR, where the true degradation kernels are unknown \cite{cornillere2019blind, gu2019ikc}.

\begin{figure*}
\centering
\includegraphics[width=\linewidth]{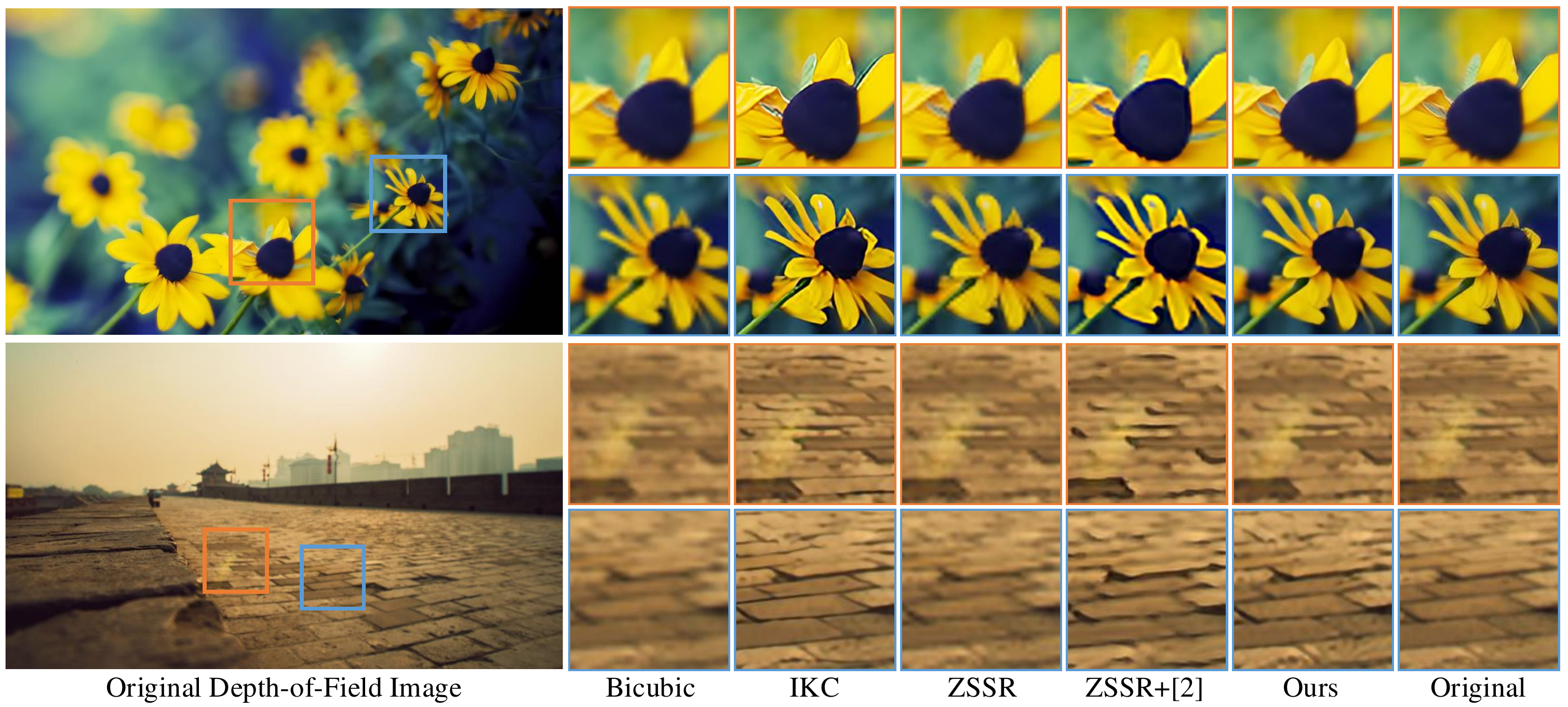}
\caption{Qualitative comparison on artistic photographs with intentional blur for $\times4$. Some methods (IKC \cite{gu2019ikc}, ZSSR \cite{shocher2018zssr}+KernelGAN \cite{bell2019kernelgan}) over-sharpen even the background (out-of-focus) regions that should be left blurry, while others generate blurry foreground (in-focus) regions. Our KOALAnet handles both regions well, generating results with similar blurriness characteristics as the original image.}
\vspace{-0.4em}
\label{fig:dof_imgs}
\end{figure*}

However, this unknown blur may be of various types with different characteristics. Often, images are captured with a different depth-of-field (DoF) by manipulating the aperture sizes and the focal lengths of camera lenses, for aesthetic purposes (e.g. Bokeh effect) as shown in Fig. \ref{fig:dof_imgs}. Recent mobile devices even try to simulate this synthetically (e.g. portrait mode) for artistic effects \cite{wadhwa2018synthetic}. Although a camera-specific degradation could be spatially-equivariant (similar to the way LR images are generated for SR), the blur generated due to DoF of the camera would be \textit{spatially-variant}, where some areas are in focus, and others are out of focus. These types of LR images are extremely challenging for SR, since ideally, the intentional blur must be left unaltered (should not be over-sharpened) to maintain the photographer's intent after SR. However, the SR results of such images are yet to be analyzed in literature.
 
In this paper, we propose a blind SR framework based on kernel-oriented adaptive local adjustment (KOALA) of SR features, called KOALAnet, by jointly learning the degradation and restoration kernels. The KOALAnet consists of two networks: a downsampling network that estimates \textit{spatially-variant} blur kernels, and an upsampling network that fuses this information by mapping the predicted degradation kernels to the feature kernel space, predicting degradation-specific local feature adjustment parameters that are applied by \textit{spatially-variant local filtering} on the SR feature maps. After training under a random anisotropic Gaussian degradation setting, our KOALAnet is able to accurately predict the underlying degradation kernels and effectively leverage this information for SR. Moreover, it demonstrates a good generalization ability on historic images containing unknown degradations compared to previous blind SR methods. We further provide comparisons on real aesthetic DoF images, and show that our KOALAnet effectively handles images with intentional blur. Our contributions are three-fold:

\begin{itemize}\itemsep1pt
 \item We propose a blind SR framework that jointly learns \textit{spatially-variant} degradation and restoration kernels. The restoration (upsampling) network leverages novel KOALA modules to adaptively adjust the SR features based on the predicted degradation kernels. The KOALA modules are \textit{extensible}, and can be inserted into any CNN architecture for image restoration tasks.
 \item We empirically show that the proposed KOALAnet outperforms the recent state-of-the-art blind SR methods for synthesized LR images obtained under randomized degradation conditions, as well as for historic LR images with unknown degradations.
 \item We first analyze SR results on images mixed with \textit{in-focus} and \textit{out-of-focus} regions, showing that our KOALAnet is able to discern intentionally blurry areas and process them accordingly, leaving the photographer's intent unchanged after SR.
\end{itemize}

\section{Related Work}

\noindent
\textbf{Single degradation SR.}\quad
Since the first CNN-based SR method by Dong \textit{et al.} \cite{dong2014learning},
highly sophisticated deep learning networks have been proposed in image SR \cite{kim2016accurate, ledig2017photo, lim2017enhanced, shi2016real, wang2018sftgan, wang2018esrgan, zhang2018image}, achieving remarkable quantitative or qualitative performance.
Especially, Wang \textit{et al.} \cite{wang2018sftgan} introduced feature-level affine transformation based on segmentation prior to generate class-specific texture in the SR result. Although these methods perform promisingly under the ideal bicubic-degraded setting, they tend to produce \textit{over-sharpened} or \textit{blurry} results if the degradations present in the test images deviate from bicubic degradation. 

\begin{figure*}
\centering
\includegraphics[width=\linewidth]{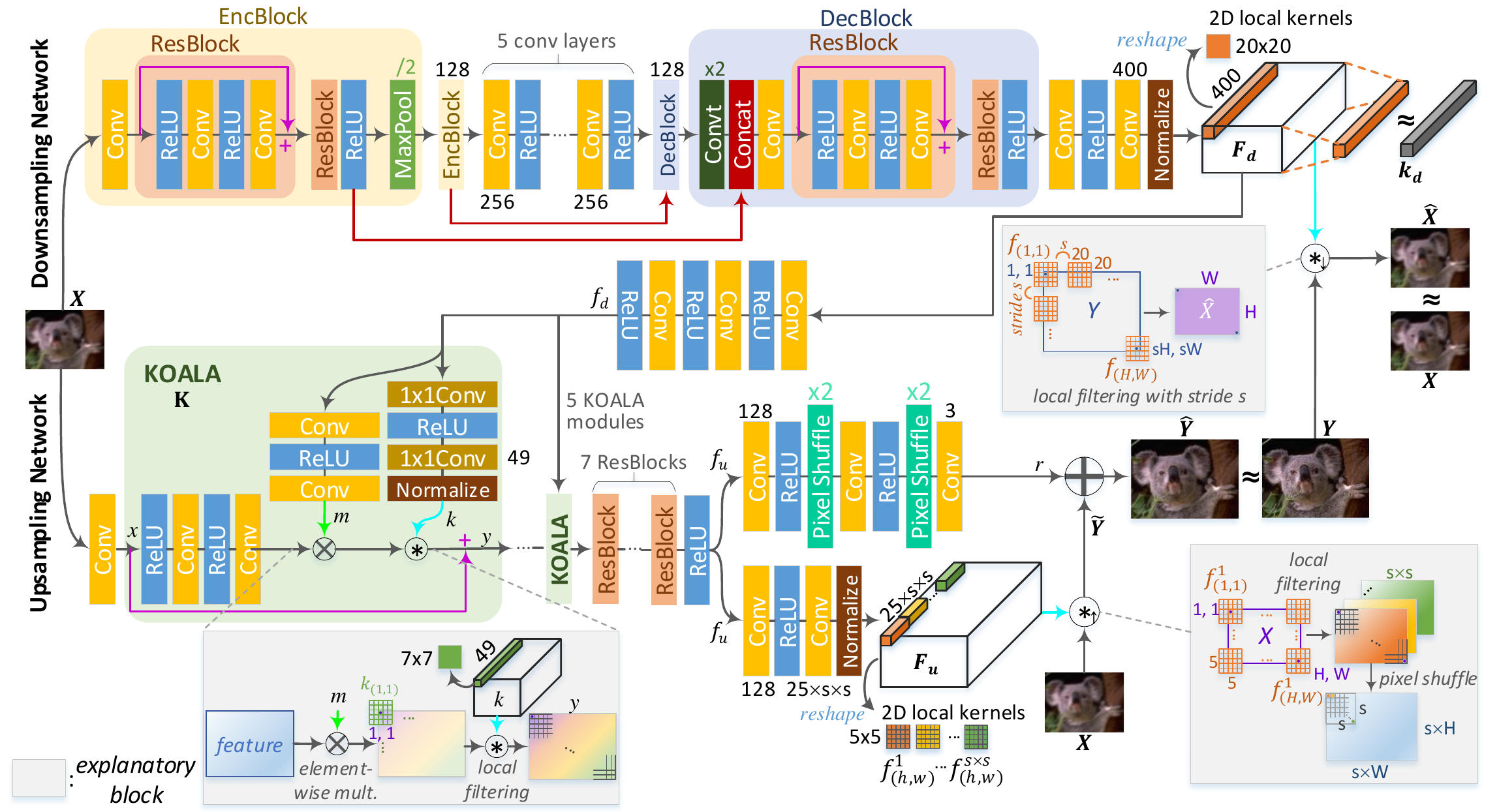}
\caption{Our proposed blind SR framework, KOALAnet, for $\times4$. The downsampling network predicts spatially-variant kernels, which are fed into the KOALA modules that in turn produce degradation-specific multiplicative (\textit{m}) and local filter parameters (\textit{k}) used to modulate the features in the upsampling network. The upsampling network generates an SR result with spatially-variant upsampling kernels.}
\label{fig:net_arch}
\vspace{-0.8em}
\end{figure*}

\smallskip\noindent
\textbf{Multiple degradation SR.}\quad
Recent methods handling multiple types of degradations can be categorized into \emph{non-blind SR} \cite{xu2020unified, zhang2018srmd}, where the LR images are coupled with the ground truth degradation information (blur kernel or noise level), or \emph{blind SR} \cite{bell2019kernelgan, cornillere2019blind, gu2019ikc}, where only the LR images are given without the ground truth degradation information that is then to be estimated. Among the former, Zhang \textit{et al.} \cite{zhang2018srmd} provided the principal components of the Gaussian blur kernel and the level of additive Gaussian noise by concatenating them with the LR input for degradation-aware SR. Xu \textit{et al.} \cite{xu2020unified} also integrated the degradation information in the same way, but with a backbone network using dynamic upsampling filters \cite{jo2018deep}, raising the SR performance. However, these methods require ground truth blur information at test time, which is unrealistic for practical application scenarios.

Among blind SR methods that predict the degradation information, an inspiring work by Gu \textit{et al.} \cite{gu2019ikc} inserted spatial feature transform modules \cite{wang2018sftgan} in the CNN architecture to integrate the degradation information with iterative kernel correction. However, the iterative framework can be time-consuming since the entire framework must be repeated many times during inference, and the optimal number of iteration loops varies among input images, requiring human intervention for maximal performance. Furthermore, their network generates vector kernels that are eventually stretched with repeated values to be inserted to the SR network, limiting the degradation modeling capability of local characteristics. 
Another prominent work is the KernelGAN \cite{bell2019kernelgan} that generates downscaled LR images by learning the internal patch distribution of the test LR image. The downscaled LR patches, or the kernel information, and the original test LR images are then plugged into zero-shot SR \cite{shocher2018zssr} or non-blind SR \cite{zhang2018srmd} methods. There also exist methods that employ GANs to generate realistic kernels for data augmentation \cite{zhou2019kernel}, or learn to synthesize LR images along with the SR image \cite{xi2020zero, guo2020closedloop}.  In comparison, our downsampling network predicts the underlying \textit{spatially-variant} blur kernels that are used to modulate and locally filter the upsampling features.

\smallskip\noindent
\textbf{Dynamic filter generation.}\quad
Jia \textit{et al.} \cite{jia2016dynamic} first proposed \textit{dynamic filter networks} that generate image- and location-specific \textit{dynamic} filters that filter the input images in a locally adaptive manner to better handle the non-stationary property of natural images, in contrast to the conventional convolution layers with spatially-equivariant filters. Application-wise, Niklaus \textit{et al.} \cite{niklaus2017videoa, niklaus2017videob} and Jo \textit{et al.} \cite{jo2018deep} successfully employed the dynamic filtering networks for video frame interpolation, and video SR, respectively. The recent non-blind SR method by Xu \textit{et al.} \cite{xu2020unified} also employed a two-branch dynamic upsampling architecture \cite{jo2018deep}. However, the provided ground truth degradation kernel is still restricted to spatially \textit{uniform} kernels and are entered naively by simple concatenation, unlike our proposed KOALAnet that estimates the underlying \textit{non-uniform} blur kernels from input LR images and effectively integrates this information for SR.

\section{Proposed Method}
We propose a blind SR framework with (i) a downsampling network that predicts spatially-variant degradation kernels, and (ii) an upsampling network that contains KOALA modules, which adaptively fuses the degradation kernel information for enhanced blind SR.
\subsection{Downsampling Network}
\label{sec:downsampler}

\noindent
\textbf{Data generation.}\quad
During training, an LR image, $X$, is generated by applying a random anisotropic Gaussian blur kernel, $k_{g}$, on an HR image, $Y$, and downsampling it with the bicubic kernel, $k_{b}$, similar to \cite{gu2019ikc, xu2020unified, zhang2018srmd}, given as,
\begin{align} 
	X = (Y*k_{g})*k_{b}\downarrow_s = (Y*k_{d})\downarrow_s,
\label{eq:degradation}
\end{align}
where $\downarrow_s$ denotes downsampling by scale factor $s$. Hence, the downsampling kernel $k_{d}$ can be obtained as $k_{d}=k_{g}*k_{b}$, and the degradation process can be implemented by an $s$-stride convolution of $Y$ by $k_{d}$. We believe that anisotropic Gaussian kernels are a more suitable choice than isotropic Gaussian kernels for blind SR, as anisotropic kernels are the more generalized superset. We do not apply any additional anti-aliasing measures (like in the default Matlab imresize function), since $Y$ is already low-pass filtered by $k_g$.

\smallskip
The downsampling network, shown in the upper part of Fig. \ref{fig:net_arch}, takes a degraded LR RGB image, $X$, as input, and aims to predict its underlying degradation kernel that is assumed to have been used to obtain $X$ from its HR counterpart, $Y$, through a U-Net-based \cite{ronneberger2015unet} architecture with ResBlocks. The output, $F_d$, is a 3D tensor of size $H\times W\times 400$, composed of $20\times20$ local filters at every $(h, w)$ pixel location. The local filters are normalized to have a sum of 1 (denoted as \textit{Normalize} in Fig. \ref{fig:net_arch}) by subtracting each of their mean values and adding a bias of $1/400$. With $F_d$, the LR image, $\hat{X}$, can be reconstructed by, 
\begin{align} 
	\hat{X} = (Y\oast F_d)\downarrow_s,
\label{eq:lr_hat}
\end{align}
where $\oast\downarrow_s$ represents $20\times20$ local filtering \cite{jia2016dynamic} at each pixel location with stride $s$, as illustrated in Fig. \ref{fig:net_arch}.

For training, we propose to use an LR reconstruction loss, $L_r=l_1(\hat{X}, X)$, which indirectly enforces the downsampling network to predict a \textit{spatially-variant degradation kernel} at each pixel location based on image prior. To bring flexibility in the spatially-variant kernel estimation, the loss with the ground truth kernel is only given to the spatial-wise mean of $F_d$. Then, the total loss for the downsampling network is given as,
\begin{align} 
	L_d=l_1(\hat{X}, X)+l_1(E_{hw}[F_d], k_d),
\label{eq:total_loss}
\end{align}
where $E_{hw}[\cdot]$ denotes a spatial-wise mean over $(h, w)$, and $k_d$ is reshaped to $1\times1\times400$ from the original size of $20\times 20$.

Estimating the blur kernel for a smooth region in an LR image is difficult since dissimilar blur kernels may produce similar smooth pixel values. Consequently, if the network aims to directly predict the true blur kernel, the gradient of a kernel matching loss may not back-propagate a desirable signal. Meanwhile, for highly textured regions of HR images, the induced LR images are largely influenced by the blur kernels, which enables the downsampling network to find inherent degradation cues from the LR images. In this case, the degradation information can be highly helpful in reconstructing the SR image as well, since most of the SR reconstruction error tends to occur in these regions. 

\subsection{Upsampling Network}
We consider the upsampling process to be the inverse of the downsampling process, and thus, design an upsampling network in correspondance with the downsampling network as shown in Fig. \ref{fig:net_arch}. The upsampling network takes in the degraded LR input, $X$, of size $H\times W\times 3$, and generates an SR output, $\hat{Y}$, of size $sH\times sW\times 3$, where $s$ is a scale factor. In the early convolution layers of the upsampling network, the SR feature maps are adjusted by five cascaded KOALA modules, K, which are explained in detail in the next section. 
Then, after seven cascaded residual blocks, R, the resulting feature map, $f_u$, is given by, $f_u = (\text{RL}\circ \text{R}^7\circ \text{K}^5\circ \text{Conv})(X),$
where RL is ReLU activation \cite{glorot2011deep}. $f_u$ is fed separately into a residual branch and a filter generation branch similar to \cite{jo2018deep}, where the residual map, $r$, and local upsampling filters, $F_u$, are obtained as,
\begin{align} 
	& r = (\text{Conv}\circ \text{PS}\circ \text{RL}\circ \text{Conv}\circ \text{PS}\circ \text{RL}\circ \text{Conv})(f_u),\label{eq:2}\\
	& F_u = (\text{Normalize}\circ \text{Conv}\circ \text{RL}\circ \text{Conv})(f_u),
\label{eq:3}
\end{align}
for $s=4$, where PS is a pixel shuffler \cite{shi2016real} of $s=2$ and Normalize denotes normalizing by subtracting the mean and adding a bias of $1/25$ for each $5\times 5$ local filter. The second PS and its preceding convolution layer are removed when generating $r$ for $s=2$. 

When applying the generated $F_u$ of size $H\times W\times (25\times s\times s)$ on the input $X$, $F_u$ is split into $s\times s$ tensors in the channel direction, and each chunk of $H\times W\times 25$ tensor is interpreted as a $5\times 5$ local filter at every $(h, w)$ pixel location. They are applied on $X$ (same filters for RGB channels) by computing the local inner product at the corresponding grid position $(h, w)$. After filtering all of the $s\times s$ chunks, the produced $H\times W\times (s\times s\times 3)$ tensor is pixel-shuffled with scale $s$ to generate the enlarged $\tilde{Y}$ of size $sH\times sW\times 3$ similar to \cite{jo2018deep}. Finally, $\hat{Y}$ is computed as $\hat{Y}=\tilde{Y}+r$, and the upsampling network is trained with $l_1(\hat{Y}, Y)$.

\begin{table*}
    \begin{center}
    \scalebox{0.8}{
    \begin{tabular}{ccccccccc}
    \toprule
    \multirow{2}{*}{\textbf{Method $\times2$}} & \textbf{Set5} & \textbf{Set14} & \textbf{BSD100} & \textbf{Urban100} & \textbf{Manga109} & \textbf{DIV2K-val} & \textbf{DIV2KRK\cite{bell2019kernelgan}} & \textbf{Complexity}\\
    & PSNR/SSIM & PSNR/SSIM & PSNR/SSIM & PSNR/SSIM & PSNR/SSIM & PSNR/SSIM & PSNR/SSIM & Time (s)/GFLOPs\\
    \midrule
    Bicubic & 27.11/0.7850 & 26.00/0.7222 & 26.09/0.6838 & 22.82/0.6537 & 24.87/0.7911 & 28.27/0.7835 & 28.73/0.8040 & - / -\\
    ZSSR \cite{shocher2018zssr} & 27.30/0.7952 & 26.55/0.7402 & 26.46/0.7020 & 23.13/0.6706 & 25.43/0.8041 & 28.69/0.7958 & 29.10/0.8215 & 24.69/\underline{6,238}\\
    KernelGAN \cite{bell2019kernelgan} & \multirow{2}{*}{27.35/0.7839} & \multirow{2}{*}{24.57/0.7061} & \multirow{2}{*}{25.56/0.6990} & \multirow{2}{*}{23.12/0.6907} & \multirow{2}{*}{25.99/0.8270} & \multirow{2}{*}{27.66/0.7892} & \multirow{2}{*}{\underline{30.36}/\underline{0.8669}} & \multirow{2}{*}{230.66/10,219}\\
    +ZSSR \cite{shocher2018zssr} & & & & & & & &\\
    BlindSR \cite{cornillere2019blind} & \underline{28.61}/\underline{0.8371} & \underline{26.63}/\underline{0.7686} & \underline{26.86}/\underline{0.7381} & \underline{24.11}/\underline{0.7396} & \underline{26.19}/\underline{0.8499} & \underline{28.90}/\underline{0.8227} & 29.44/0.8464 & \underline{9.21}/13,910\\
    \midrule
    \textbf{Ours} & \textbf{33.08}/\textbf{0.9137} & \textbf{30.35}/\textbf{0.8568} & \textbf{29.70}/\textbf{0.8248} & \textbf{27.19}/\textbf{0.8318} & \textbf{32.61}/\textbf{0.9369} & \textbf{32.55}/\textbf{0.8902} & \textbf{31.89}/\textbf{0.8852} & \textbf{0.71}/\textbf{201}\\
    \bottomrule
    \toprule
    \textbf{Method $\times4$} & \textbf{Set5} & \textbf{Set14} & \textbf{BSD100} & \textbf{Urban100} & \textbf{Manga109} & \textbf{DIV2K-val} & \textbf{DIV2KRK\cite{bell2019kernelgan}} & \textbf{Complexity}\\
    \midrule
    Bicubic & 26.41/0.7511 & 24.73/0.6641 & 25.12/0.6321 & 22.04/0.6061 & 23.60/0.7482 & 27.04/0.7417 & 25.33/0.6795 & - / -\\
    ZSSR \cite{shocher2018zssr} & 26.49/0.7530 & 24.93/0.6812 & 25.36/0.6526 & 22.39/0.6327 & 24.43/0.7813 & 27.39/0.7590 & 25.61/0.6911 & 16.91/6,091\\
    KernelGAN \cite{bell2019kernelgan} & \multirow{2}{*}{22.12/0.5989} & \multirow{2}{*}{19.73/0.5194} & \multirow{2}{*}{21.02/0.5377} & \multirow{2}{*}{20.12/0.5743} & \multirow{2}{*}{22.61/0.7345} & \multirow{2}{*}{23.75/0.6830} & \multirow{2}{*}{26.81/0.7316} & \multirow{2}{*}{357.70/11,908}\\
    +ZSSR \cite{shocher2018zssr} & & & & & & & &\\
    IKC \cite{gu2019ikc} \textit{last} & 27.73/0.8024 & 25.38/0.7162 & 25.68/0.6844 & 23.03/0.6852 & 25.44/0.8273 & 27.61/0.7843 & 27.39/\underline{0.7639} & \underline{0.67}/\underline{575}\\
    IKC \cite{gu2019ikc} \textit{max} & \underline{28.04}/\underline{0.8079} & \underline{25.85}/\underline{0.7261} & \underline{26.01}/\underline{0.6951} & \underline{23.21}/\underline{0.6943} & \underline{25.82}/\underline{0.8361} & \underline{27.98}/\underline{0.7912} & \underline{27.70}/\textbf{0.7684} & \underline{0.67}/\underline{575}\\
    \midrule
    \textbf{Ours} & \textbf{30.28}/\textbf{0.8658} & \textbf{27.20}/\textbf{0.7541} & \textbf{26.97}/\textbf{0.7172} & \textbf{24.71}/\textbf{0.7427} & \textbf{28.48}/\textbf{0.8814} & \textbf{29.44}/\textbf{0.8156} & \textbf{27.77}/0.7637 & \textbf{0.59}/\textbf{57}\\
    \bottomrule
    \end{tabular}}
    \caption{Quantitative comparison on various datasets. We also provide a comparison on computational complexity in terms of the average inference time on Set5, and GFLOPs on ``\textit{baby}'' in Set5. \textbf{Bold} and \underline{underline} indicate the best and the second best performance, respectively.} \label{table: quant_compar}
    \end{center}
    \vspace{-1.3em}
\end{table*}

\smallskip\noindent
\textbf{Kernel-oriented adaptive local adjustment.}\quad
\label{KOALA}
We propose a novel feature transformation module, KOALA, that adaptively adjusts the intermediate features in the upsampling network based on the degradation kernels predicted by the downsampling network. The KOALA modules are placed at the earlier stage of feature extraction in order to calibrate the anisotropically degraded LR features before the reconstruction phase. 

Specifically, when the input feature, $x$, is entered into a KOALA module, K, it goes through 2 convolution layers, and is adjusted by a set of multiplicative parameters, $m$, followed by a set of local kernels, $k$, generated based on the predicted degradation kernels, $F_d$. Instead of directly feeding $F_d$ into K, the kernel features, $f_d$, extracted after 3 convolution layers are entered. After a local residual connection, the output, $y$, of the KOALA module is given by,
\begin{align} 
	y &= \text{K}(x\mid f_d) \notag\\
	&= \{(\text{Conv}\circ \text{RL}\circ \text{Conv}\circ \text{RL})(x)\otimes m\}\oast k + x, \label{eq:4}
\end{align}
where,
\vspace{-0.5em}
\begin{align}
	m &= (\text{Conv}\circ \text{RL}\circ \text{Conv})(f_d),\quad \text{and} \\
	k &= (\text{Normalize}\circ \text{Conv}_{1\times1}\circ \text{RL}\circ \text{Conv}_{1\times1})(f_d).
	\label{eq:5}
\end{align}
In Eq. \ref{eq:4}, $\otimes$ and $\oast$ denote element-wise multiplication and local feature filtering, respectively. For generating $k$, $1\times1$ convolutions are employed so that spatially adjacent values of kernel features, $f_d$, are not mixed by convolution operations. The kernel values of $k$ are constrained to have a sum of 1 (Normalize), like for $F_d$ and $F_u$.

The local feature filtering operation, $\oast$, is first applied by reshaping a $1\times 1\times 49$ vector at each grid position $(h, w)$ to a $7\times 7$ 2D local kernel, and computing the local inner product at each $(h, w)$ position of the input feature. Since the same $7\times 7$ kernels are applied channel-wise, the multiplicative parameter, $m$, introduces element-wise scaling for the features over the channel depth. This is also efficient in terms of the number of parameters, compared to predicting the per-pixel local kernels for every channel ($49+64$ vs. $49\times64$ filter parameters). By placing the residual connection after the feature transformations (Eq. \ref{eq:4}), the adjustment parameters can be considered as removing the unwanted feature residuals related to degradation from the original input features.

\subsection{Training Strategy} 
We employ a 3-stage training strategy: (i) the downsampling network is pre-trained with $l_1(\hat{X}, X)$; (ii) the upsampling network is pre-trained with $l_1(\hat{Y}, Y)$ by replacing all KOALA modules with ResBlocks; (iii) the whole framework (KOALAnet) including the KOALA modules (with convolution layers needed for generating $f_d$, $m$ and $k$ inserted on the pre-trained ResBlocks) is jointly optimized based on $l_1(\hat{X}, X)+l_1(\hat{Y}, Y)$. With this strategy, the KOALA modules can be effectively trained with already meaningful features obtained from the early training phases, and focus on utilizing the degradation kernel cues for SR.

\section{Experiment Results}
\noindent
\textbf{Data generation.}\quad
In our implementations, $k_{d}$ of size $20\times20$ is computed by convolving $k_{b}$ with a random anisotropic Gaussian kernel ($k_{g}$) of size $15\times15$, following Eq. \ref{eq:degradation}. It should be noted that $k_{b}$ is originally a bicubic downscaling kernel of size $4\times4$ same as in the \textit{imresize} function of Matlab \textit{without} anti-aliasing, but is zero-padded to be $20\times20$ to align with the size of $k_{d}$ as well as to avoid image shift. The Gaussian kernels for degradation are generated by randomly rotating a bivariate Gaussian kernel by $\theta\sim \text{Uniform}(0, \pi/2)$, and by randomly selecting its kernel width that is determined by a diagonal covariance matrix with $\sigma_{11}$ and $\sigma_{22} \sim \text{Uniform}(0.2,4.0)$. With $k_{d}$, we build our training data on the DIV2K \cite{DIV2K} dataset according to Eq. \ref{eq:degradation}. Testsets are generated using Set5 \cite{Set5}, Set14 \cite{Set14}, BSD100 \cite{BSD100}, Urban100 \cite{Urban100}, Manga109 \cite{Manga109} and DIV2K-val \cite{DIV2K} for comparison with other methods. When generating the testsets, we ensure that different parameters are selected for different images by assigning different random seed values. We additionally compare to DIV2KRK proposed in \cite{bell2019kernelgan}, which contains DIV2K images that are randomly degraded.

\begin{figure*}
\captionsetup[subfigure]{aboveskip=0.5mm}
\centering
\begin{subfigure}[t]{0.497\linewidth}  
\includegraphics[width=\linewidth]{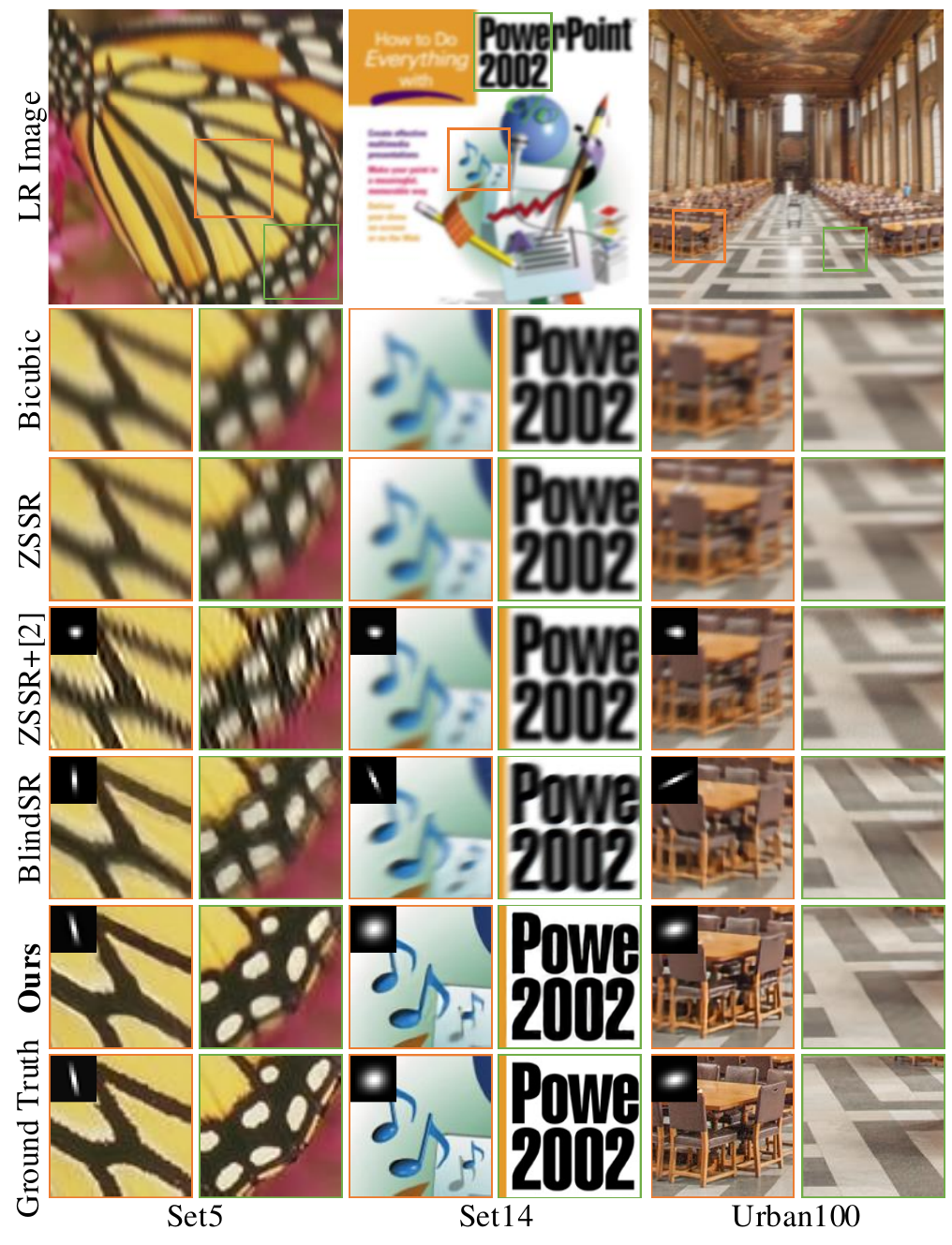}
\caption{Scale factor 2}
\end{subfigure}
\begin{subfigure}[t]{0.497\linewidth}
\includegraphics[width=\linewidth]{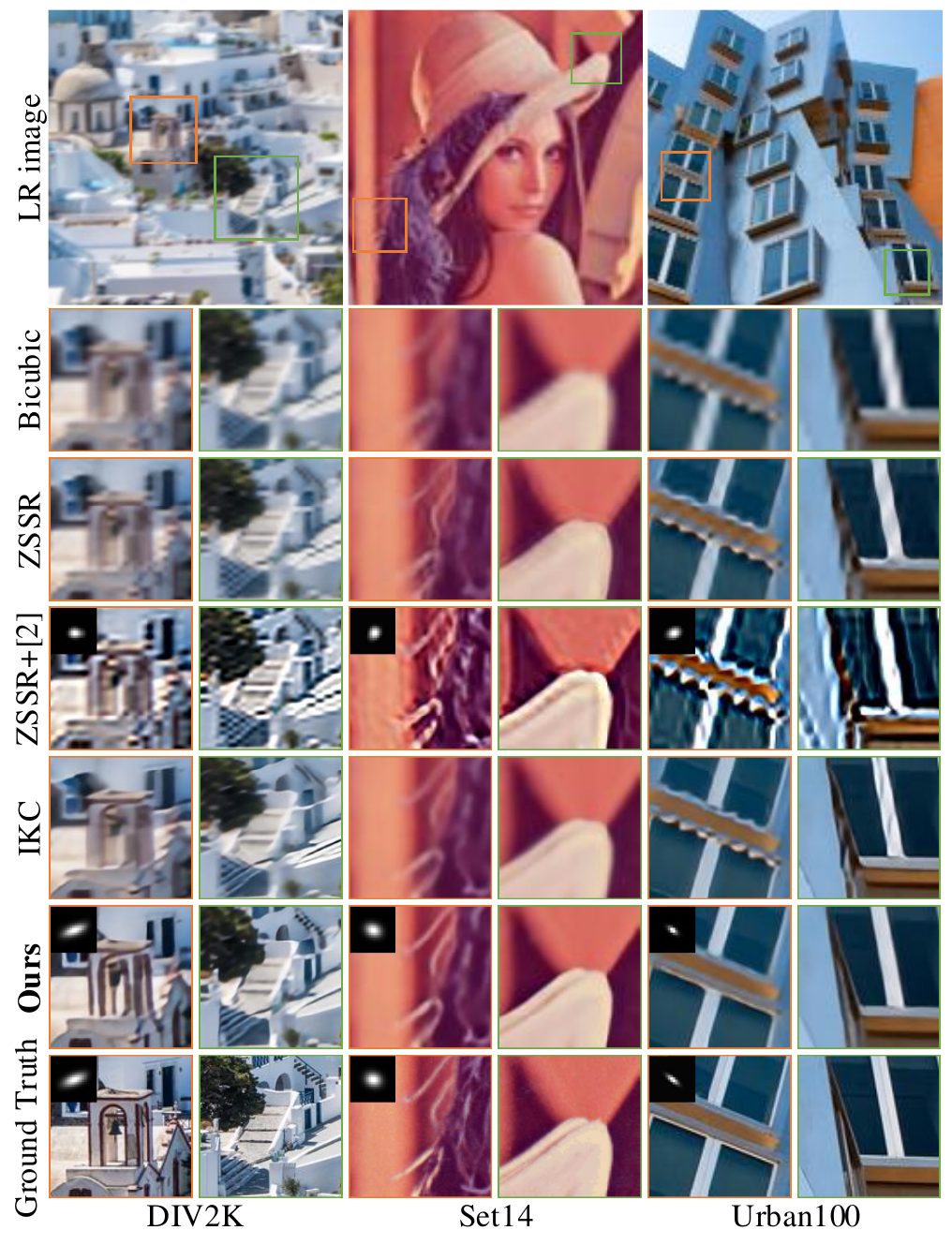}
\caption{Scale factor 4}
\end{subfigure}
\vspace{-0.2em}
\caption{Qualitative comparison to other methods. The estimated or ground truth degradation kernels are placed on the top left.}
\label{fig:qual_comp}
\vspace{-0.5em}
\end{figure*}

\smallskip\noindent
\textbf{Training parameters.}\quad
All convolution filters in the KOALAnet are of size $3\times 3$ with 64 output channels following \cite{kim2016accurate}, unless otherwise noted as $1\times 1$Conv or with the output channel noted next to an operation block in Fig. \ref{fig:net_arch}. All CNN-based networks used in our experiments are trained with LR patches of size $64\times 64$ normalized to $[-1, 1]$, where each patch is randomly cropped, and randomly degraded with $k_d$ during training. The mini-batch size is 8, and the initial learning rate of $10^{-4}$ is decreased by $1/10$ at 80\% and 90\% of 200K iterations for each training stage. We consider $s=2$ and $s=4$ for SR in our experiments.

\subsection{Comparison to Existing Blind SR Methods}
We compare our method with recent state-of-the-art blind SR methods, BlindSR \cite{cornillere2019blind} and IKC \cite{gu2019ikc}. For \cite{cornillere2019blind}, we use the pre-trained model in an independent implementation by an author with only $s=2$ model. For \cite{gu2019ikc}, we use the official pre-trained model by the authors with only $s=4$ model. We also compare against ZSSR \cite{shocher2018zssr} with default degradation as well as by incorporating KernelGAN \cite{bell2019kernelgan} to provide degradation information, both with the official codes.

\smallskip\noindent
\textbf{Quantitative comparison.}\quad
We compare the Y-channel PSNR and SSIM for various methods on the six random anisotropic degradation testsets as well as DIV2KRK \cite{bell2019kernelgan} in Table \ref{table: quant_compar}. We also provide a comparison on the average inference time and GFLOPs in the rightmost column, where the inference time in seconds is measured on Set5 with an NVIDIA Titan RTX excluding file I/O times, and the GFLOPs is computed on the ``\textit{baby}'' image in Set5 that is of $512\times512$ resolution in terms of the HR ground truth. The inference time includes zero-shot training for ZSSR \cite{shocher2018zssr} and optimization for \cite{bell2019kernelgan, cornillere2019blind}. The inherent limitation of zero-shot models is that they cannot leverage the abundant training data that is utilized by other methods as image-specific CNNs are trained at test time. For IKC \cite{gu2019ikc}, we report the results of the last iteration (IKC \textit{last}) as well as those producing the maximum PSNR (IKC \textit{max}) from total 7 iteration loops. Since IKC is trained under an isotropic setting, its data modeling capability tends to be limited under a more randomized superset of anisotropic blur. Note that BlindSR \cite{cornillere2019blind} had been trained under an anisotropic setting like ours. On DIV2KRK \cite{bell2019kernelgan}, where artificial noise is injected to the synthetic degradation kernels, internal-learning-based methods ZSSR \cite{shocher2018zssr} and KernelGAN \cite{bell2019kernelgan} are advantageous over the other methods as they can adapt to the unorthodox kernels. Nevertheless, our method outperforms all compared methods on DIV2KRK even though it was not trained with kernel noise, demonstrating good generalization ability. On other testsets, our KOALAnet outperforms the other methods by a large margin of over 1 dB in most cases.

\begin{figure}
\centering
\includegraphics[width=\columnwidth]{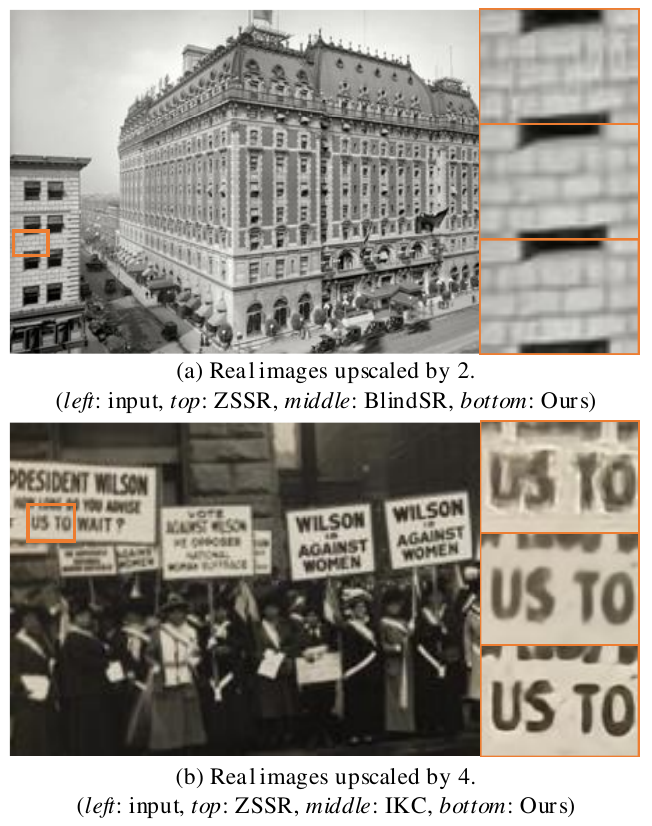}
\caption{Qualitative results on old historic images \cite{LapSRN}.}
\vspace{-0.5mm}
\label{fig:real_imgs}
\end{figure}

\smallskip\noindent
\textbf{Qualitative comparison.}\quad
We compare the visual results on the randomized anisotropic testset in Fig. \ref{fig:qual_comp}. We have also visualized the mean of the predicted spatially-variant kernels along with the ground truth kernels. Our method is able to restore sharp edges and high frequency details. Most importantly, in Fig. \ref{fig:real_imgs}, we also compare our method under real conditions on old historic images \cite{LapSRN} without ground truth labels. In this case, the results are generated using the configuration for real images for ZSSR, and we show the results generated at the last iteration for IKC. Our method performs well even on these real images with unknown degradations, demonstrating good generalization ability.

\subsection{Results on Aesthetic Images}
\begin{figure}
\centering
\includegraphics[width=\columnwidth]{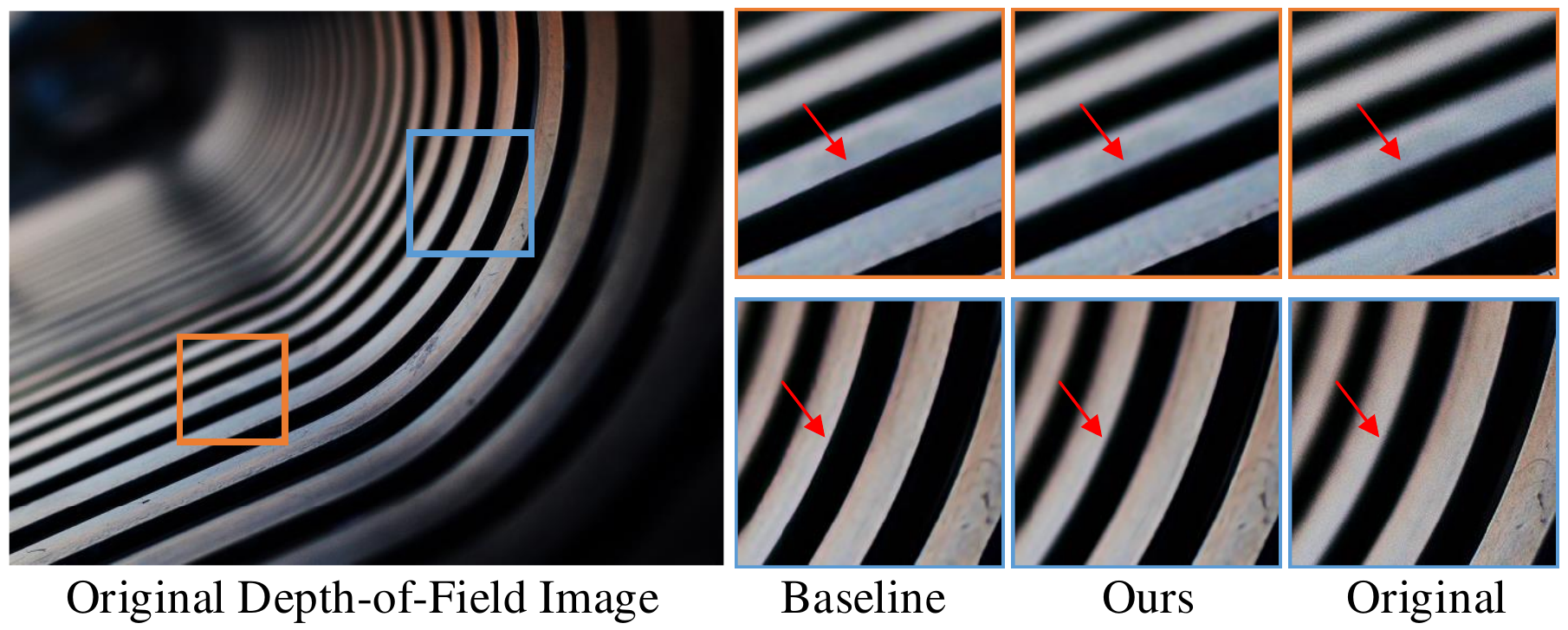}
\caption{Comparison with the Baseline on a shallow DoF image for $\times4$. The Baseline with only the upsampling network tends to generate a deeper DoF image compared to the original image.}
\label{fig:dof_imgs_baseline}
\end{figure}

We collected several shallow DoF images from the web containing intentional spatially-variant blur for aesthetic purposes, to compare the SR results of existing blind SR methods \cite{bell2019kernelgan, gu2019ikc, shocher2018zssr} to ours. Before applying SR, these images are bicubic-downsampled so that we can consider the original images as ground truth, in order to gauge the intended blur characteristics in the original images. As shown in Fig. \ref{fig:dof_imgs}, IKC \cite{gu2019ikc} and ZSSR \cite{shocher2018zssr} with KernelGAN \cite{bell2019kernelgan} tend to over-sharpen even the intentionally blurry areas that should be left blurry. ZSSR \cite{shocher2018zssr} produces blurry results overall, even in the foreground (in-focus) regions. In contrast, our KOALAnet leaves the originally out-of-focus region blurry and appropriately upscales the overall image, yielding results that are closest to the original images. For further analysis, we also compare our KOALAnet with a Baseline that only has the upsampling network in our framework. As shown in Fig. \ref{fig:dof_imgs_baseline}, the regions with strong blur far away from the in-focus area remain blurry for both methods. However, the Baseline cannot correctly disentangle the intentional blur from the degradation blur in the boundary areas between the in-focus and completely out-of-focus areas, where it can be ambiguous whether the blur should be sharpened or left blurry. With shallow DoF images where only a narrow band of regions are in focus in Fig. \ref{fig:dof_imgs_baseline}, the Baseline tends to produce results with a \textit{deeper DoF} than the original image due to over-sharpening of the boundary areas. 

\begin{figure}
\centering
\includegraphics[width=\columnwidth]{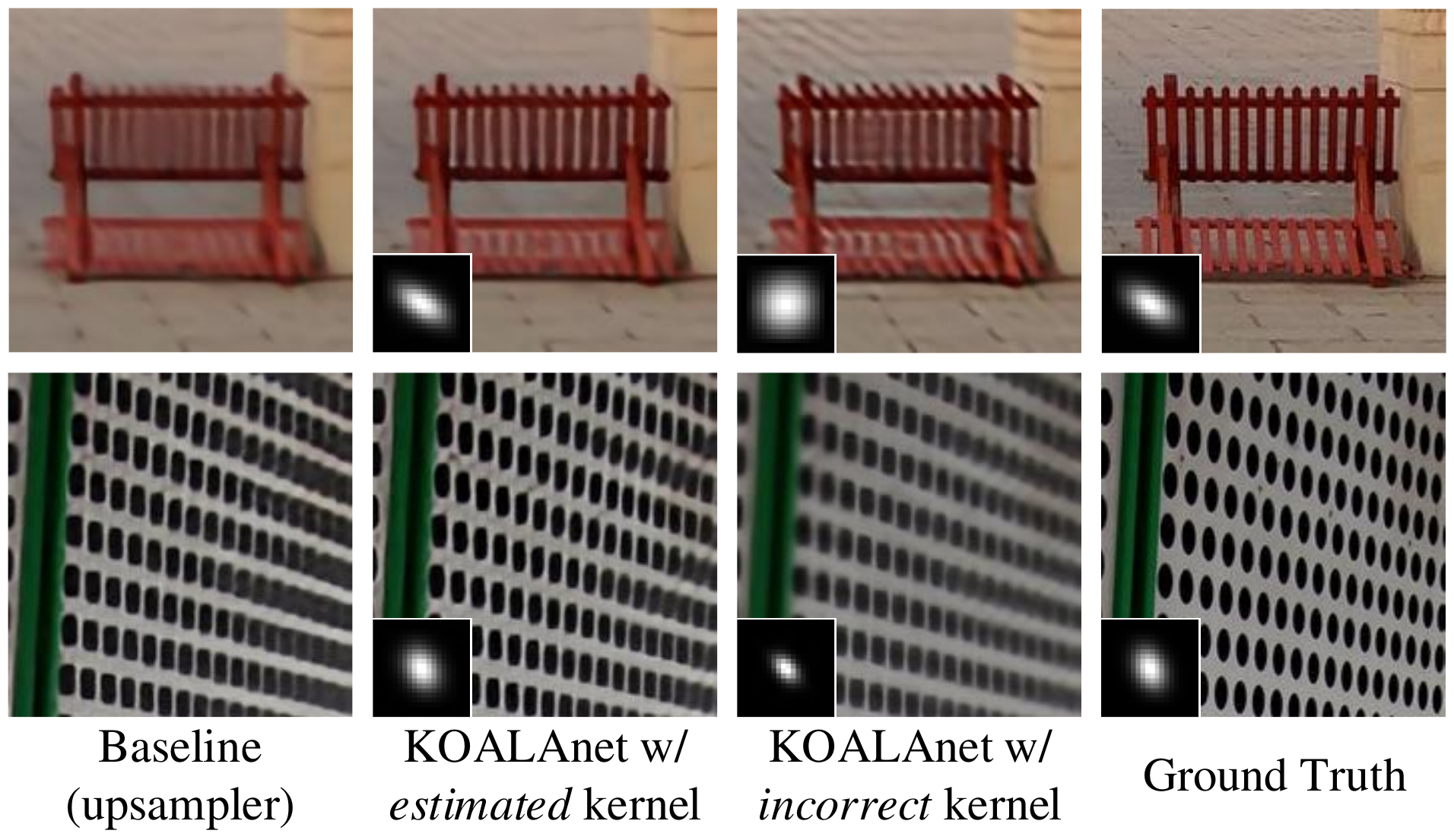}
\caption{Effect of the degradation kernels on KOALAnet for $\times4$.}
\label{fig:koala_ab_natural}
\end{figure}

\subsection{Ablation Study}
\label{architecture}
In this Section, we analyze the effect of the different components in our framework with various ablation studies, and provide visualizations of estimated blur kernels, local upsampling filters, and local filters in the KOALA modules.

\begin{table}
    \begin{center}
    \scalebox{0.85}{
    \begin{tabular}{llll|l}
    \toprule
    \multirow{2}{*}{Model} & Baseline & KOALA & \multirow{2}{*}{KOALAnet} & KOALA\\
    & (upsampler) & only \textit{k} & & +GT kernel\\
    \midrule
    PSNR & 29.20 & 29.40 & 29.44 & \textbf{29.67}\\
    SSIM & 0.8110 & 0.8150 & 0.8156 & \textbf{0.8212}\\
    \bottomrule
    \end{tabular}}
    \caption{Ablation study on the KOALA module for $\times$4.}
    \label{table: koala}
    \end{center}
\vspace{-5mm}
\end{table}

\smallskip\noindent
\textbf{Effect of KOALA modules.}\quad
We analyzed the effect of the proposed KOALA modules by retraining the following SR models: (i) Baseline with only the upsampling network without using any degradation kernel information (no downsampling network, nor KOALA modules), (ii) a model that only has the parameters \textit{k} (not \textit{m}) in the KOALA modules, (iii) a model to which ground truth kernels are given instead of the estimated degradation kernels (KOALA+GT kernel). From the Baseline, adding KOALA modules with only \textit{k} parameters improves PSNR performance by 0.2 dB, and adding \textit{m} further improves the PSNR gain by 0.04 dB, showing the effectiveness of our proposed KOALA modules in incorporating degradation kernel information. The SR performance of KOALA modules with ground truth kernels can be considered as an upper bound, with 0.23 dB higher PSNR than the KOALAnet. Fig. \ref{fig:koala_ab_natural} compares the $\times 4$ SR results of the Baseline and the KOALAnet. With the predicted degradation kernel information, the KOALA modules help to effectively remove the blur induced by the random degradations, while revealing fine edges ($2^{\text{nd}}$ column). The images in the $3^{\text{rd}}$ column show the SR results when incorrect larger or smaller blur kernels are deliberately provided to the KOALA modules. In these cases, the wrong kernels cause the upsampling network to produce over-sharpened (for larger kernels) or blurry (for smaller kernels) results. All models were tested on the DIV2K-val testset.

\begin{table}
    \begin{center}
    \scalebox{0.82}{
    \begin{tabular}{ccccc|c}
    \toprule
    \multirow{2}{*}{Models} & (a) & (b) & (c) & (d) & Full \\
    & ResNet & U-Net & Uniform & No Norm. & Model\\
    \midrule
    PSNR & 47.79 & 47.85 & 45.78 & 49.15 & \textbf{49.50}\\
    SSIM & 0.9967 & 0.9967 & 0.9945 & 0.9976 & \textbf{0.9980}\\
    \bottomrule
    \end{tabular}
    }
    \caption{Experiment on downsampling network architecture for $\times$1/4. All networks contain 27 convolution layers.}
    \label{table: downsampler}
    \end{center}
\vspace{-1mm}
\end{table}

\begin{table}
    \begin{center}
    \scalebox{0.85}{
    \begin{tabular}{cccc}
    \toprule
    $l_2$ error & KernelGAN\cite{bell2019kernelgan} & BlindSR\cite{cornillere2019blind} & KOALAnet \\
    \midrule
    DIV2K-val & 0.0230 & 0.0152 & \textbf{0.0010}\\
    DIV2KRK\cite{bell2019kernelgan} & 0.0084 & 0.0081 & \textbf{0.0044}\\
    \bottomrule
    \end{tabular}}
    \caption{Kernel accuracy (average $l_2$ error) measured on the two random anisotropic datasets, DIV2K-val and DIV2KRK \cite{bell2019kernelgan}.}
    \label{table: kernel_accuracy}
    \end{center}
\vspace{-3mm}
\end{table}

\smallskip\noindent
\textbf{Downsampling network architecture.}\quad
To analyze the downsampling network, we retrained its four variants: (a) a ResNet model -- a ResNet-style architecture trained instead of a U-Net, (b) a model with residual connection removed from ResBlocks -- thus a common U-Net, not a ResU-Net, (c) a model that estimates uniform kernels, (d) a model without normalization (sum to 1) employed for $F_d$ (\textit{No Norm}). Table \ref{table: downsampler} shows PSNR and SSIM values measured between the original degraded LR image and the LR image reconstructed using the kernels produced from the different models, on the DIV2K-val testset. \textit{Full Model} in Table \ref{table: downsampler} denotes our final downsampling network (a ResU-Net), where there is a large drop in PSNR if any of the components are ablated. Especially, if spatially-equivariant kernels are estimated as in (c), PSNR performance drops drastically by 4.72 dB, showing the importance of using spatially-variant kernels.

\smallskip\noindent
\textbf{Degradation kernel estimation accuracy.}\quad
In order to evaluate the accuracy of degradation kernel estimation, we measured the average $l_2$ distance between the ground truth kernels and the kernels estimated by KernelGAN \cite{bell2019kernelgan}, BlindSR \cite{cornillere2019blind} and the downsampler of KOALAnet, on the random anisotropic degradation testset DIV2K-val and DIV2KRK \cite{bell2019kernelgan} in Table \ref{table: kernel_accuracy}. We make sure that the centers of the estimated kernels are aligned to the center of each ground truth kernel through manual shifting. As shown in Table \ref{table: kernel_accuracy}, KOALAnet predicts more accurate degradation kernels with lower $l_2$ error compared to other kernel estimators.

\begin{figure}
\centering
\includegraphics[width=\linewidth]{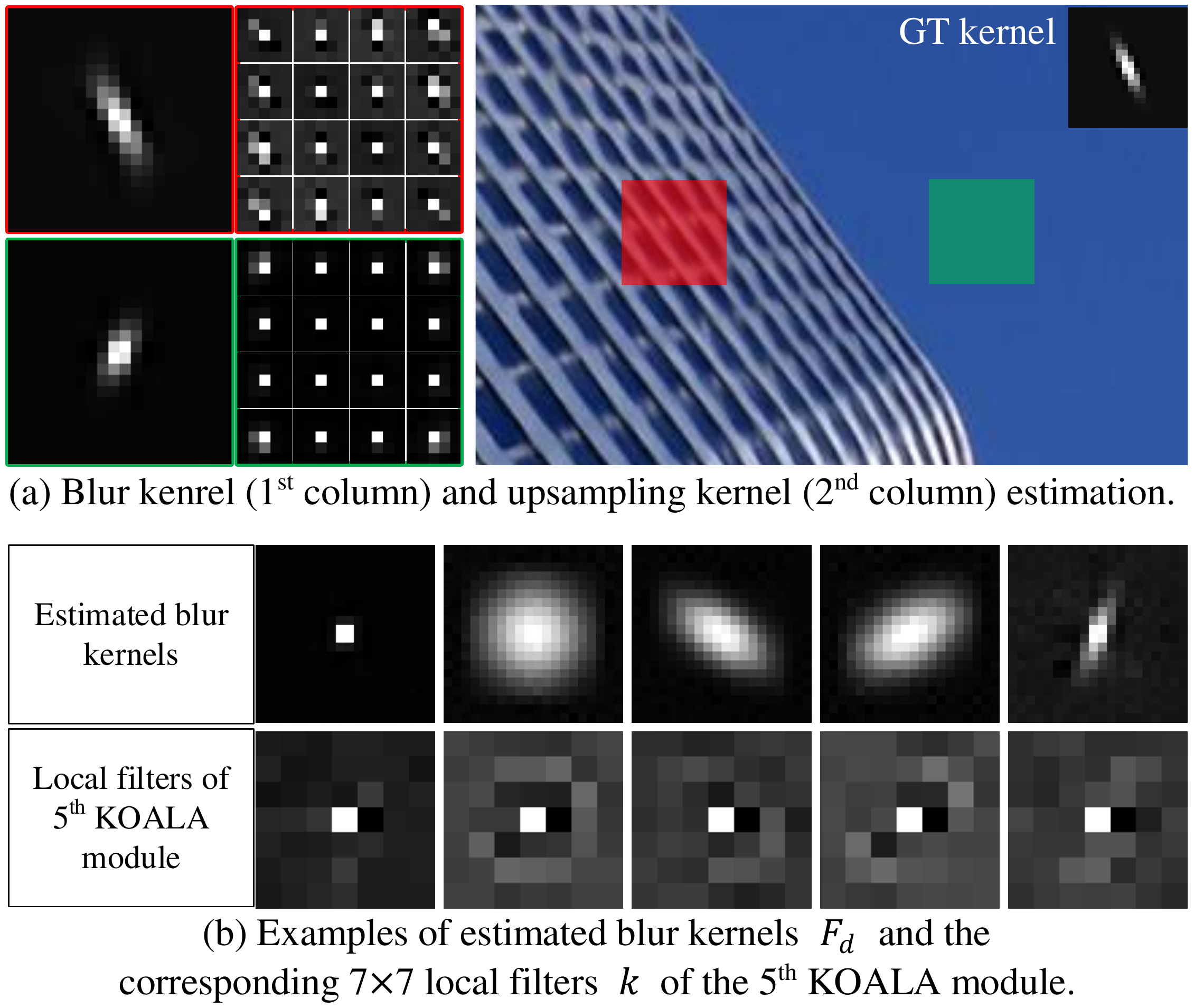}
\caption{Various kernel visualizations of KOALAnet.}
\label{fig:kernel_vis}
\end{figure}

\smallskip\noindent
\textbf{Various kernel visualizations.}\quad
In Fig. \ref{fig:kernel_vis} (a), we visualized the estimated blur kernels and upsampling kernels of our KOALAnet on two different locations in the same image. The $1^{\text{st}}$ column shows the spatially-variant degradation kernels that are predicted by the downsampling network. As discussed in Section \ref{sec:downsampler}, the predicted blur kernel is close to the true kernel in the complex area (\textit{red box}), while a non-directional kernel is obtained in the homogeneous region (\textit{green box}). In the $2^{\text{nd}}$ column, the $s\times s$ 2D upsampling kernels of size $5\times 5$ are also shown to be non-uniform depending on the location. We have also visualized some examples of local filters, $k$, of the KOALA modules in Fig. \ref{fig:kernel_vis} (b). The top row shows the degradation kernels estimated by the downsampling network, and the bottom row shows the $7\times 7$ local filters ($k$) of the $5^{\text{th}}$ KOALA module. Even without any explicit enforcement on the shape of $k$, they are learned to be related to the orientations and shapes of the blur kernel, and thus able to adjust the SR features accordingly.

\section{Conclusion}
Blind SR is an important step towards generalizing learning-based SR models for diverse types of degradations and content of LR data. In order to achieve this goal, we designed a downsampling network that predicts spatially-variant kernels and an upsampling network that leverages this information effectively, by applying these kernels as local filtering operations to modulate the early SR features based on the degradation information. As a result, our proposed KOALAnet accurately predicts the HR images under a randomized synthetic setting as well as for historic data. Furthermore, we first analyzed the SR results on real aesthetic photographs, for which our KOALAnet appropriately handles the intentional blur unlike other methods or the Baseline. Our code and data are publicly available on the web.

\small{
\smallskip\noindent
\textbf{Acknowledgement.}\quad
This work was supported by Institute for Information \& communications Technology Promotion (IITP) grant funded by the Korea government (MSIT) (No. 2017-0-00419, Intelligent High Realistic Visual Processing for Smart Broadcasting Media).
}

{\small
\bibliographystyle{ieee_fullname}

}

\clearpage
\begin{figure*}[!t]
\captionsetup[subfigure]{justification=centering}
\centering
\begin{subfigure}[t]{0.525\linewidth}
\includegraphics[width=\linewidth]{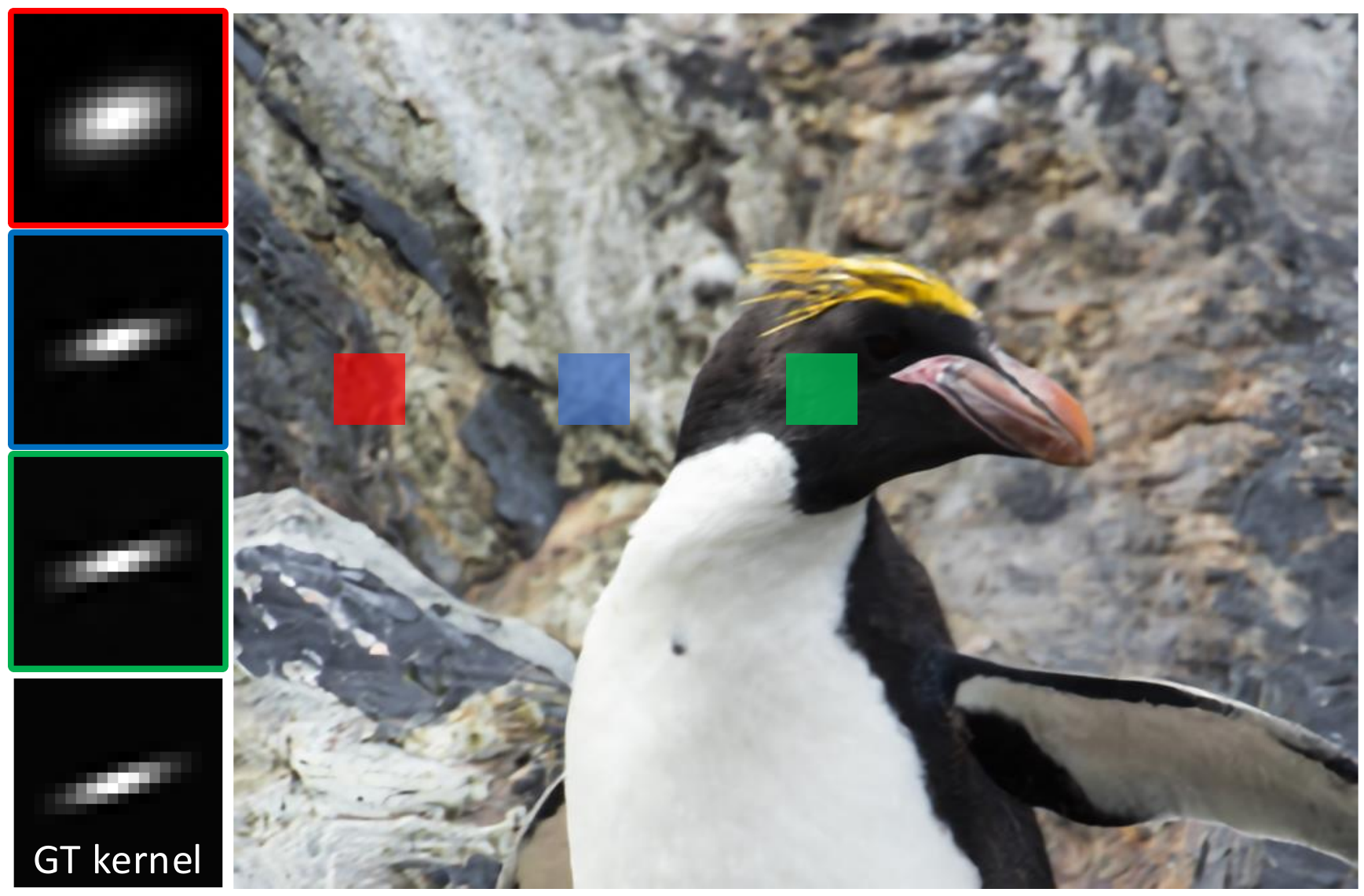}
\caption{
Estimated degradation kernels at different \\ spatial locations of the low resolution image.
}
\end{subfigure}
\begin{subfigure}[t]{0.47\linewidth}
\includegraphics[width=\linewidth]{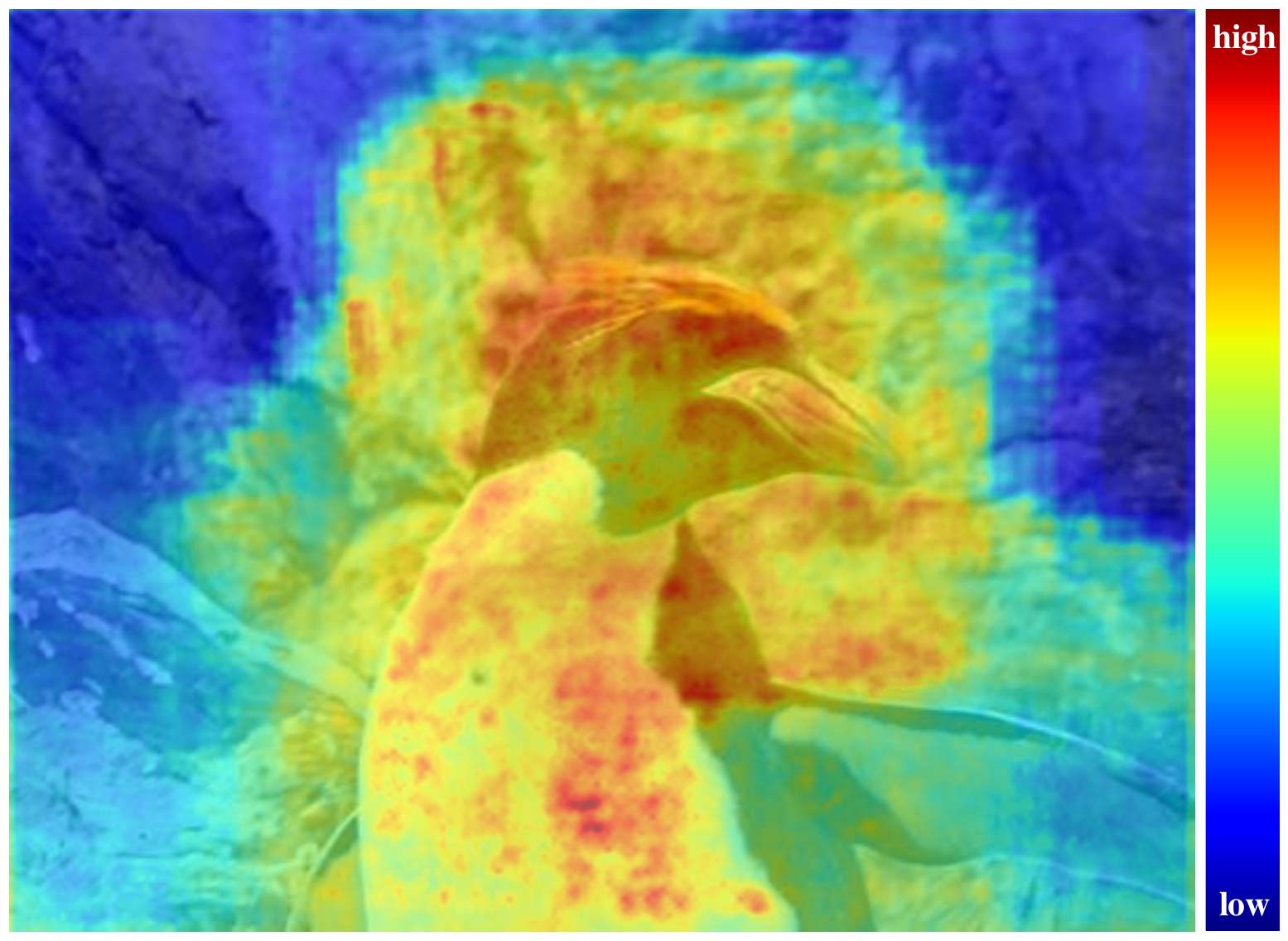}
\caption{
Visualization of the cosine similarity of the estimated per-pixel degradation kernels and the ground truth kernel.
}
\end{subfigure}
\caption{Degradation kernel visualization on different spatial locations of the same image. In this example, the penguin is the object in focus and the background is blurry (out of focus). The estimated degradation kernels differ depending on the spatial location as shown in (a). The cosine similarity of each per-pixel kernel is visualized in (b). Accurate degradation kernels are estimated near the focused region, especially in the boundary areas between the in-focus and out-of-focus areas.}
\label{fig:non_uniform_kernel_supp}
\end{figure*}

\begin{appendices}
\section{Additional Kernel Analyses}
\subsection{Spatially-variant Kernel Visualizations}
Photography enthusiasts tend to take pictures with intentionally blurry (out-of-focus) areas, in order to emphasize objects or regions of interest in the depth-of-field (DoF) by controlling the aperture size or the focal length of the camera lens to focus on the areas of interest. In the main paper, we showed that existing blind super-resolution (SR) methods tend to generate over-sharpened or blurry results for these types of artistic images (Fig. 1 in main paper). We further showed that a vanilla SR network that does not consider the degradation information over-sharpens even the \textit{intentionally} out-of-focus area, resulting in images with a deeper DoF (Fig. 5 in main paper). The over-sharpening tends to happen in the boundary regions between in-focus and completely out-of-focus areas. This is an important observation that was not previously dealt with in literature, which can be useful in handling images with intentionally blurry regions. 

\begin{figure*}
\centering
\begin{subfigure}[t]{0.98\linewidth}  
\includegraphics[width=\linewidth]{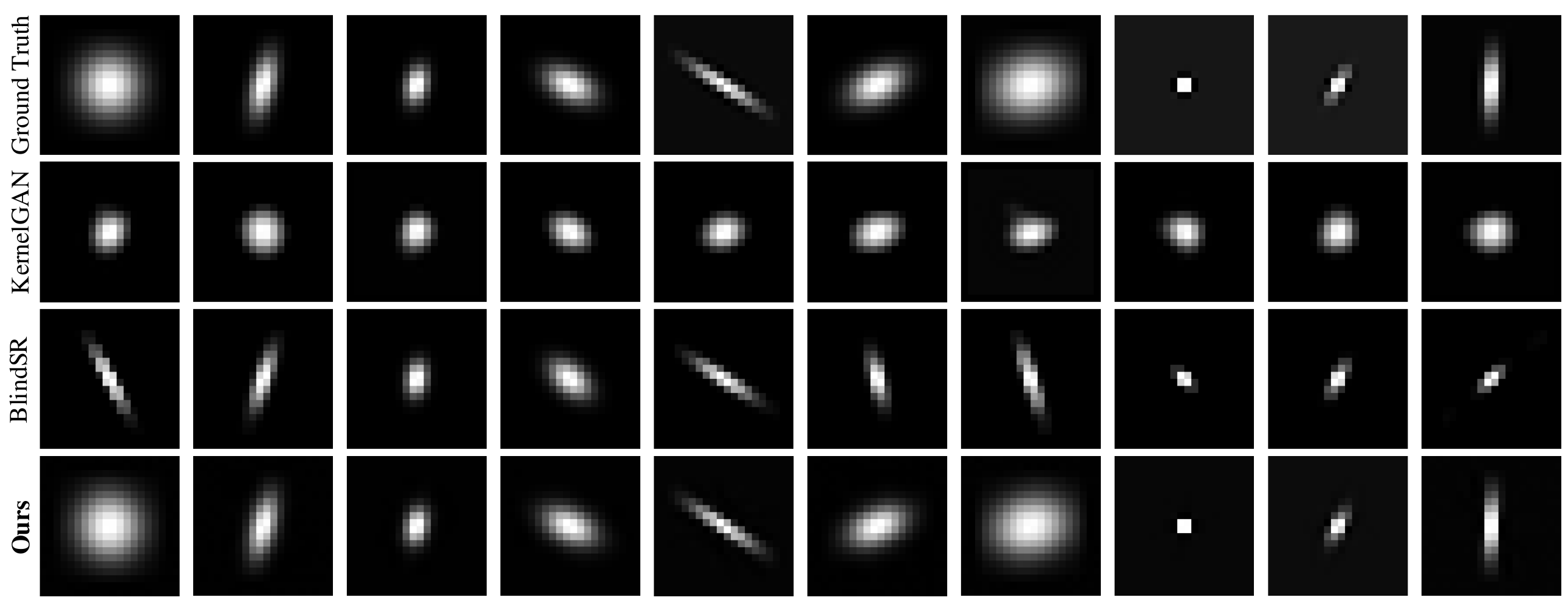}
\caption{Comparison of estimated degradation kernels on DIV2K-val}
\end{subfigure}
\begin{subfigure}[t]{0.98\linewidth}
\includegraphics[width=\linewidth]{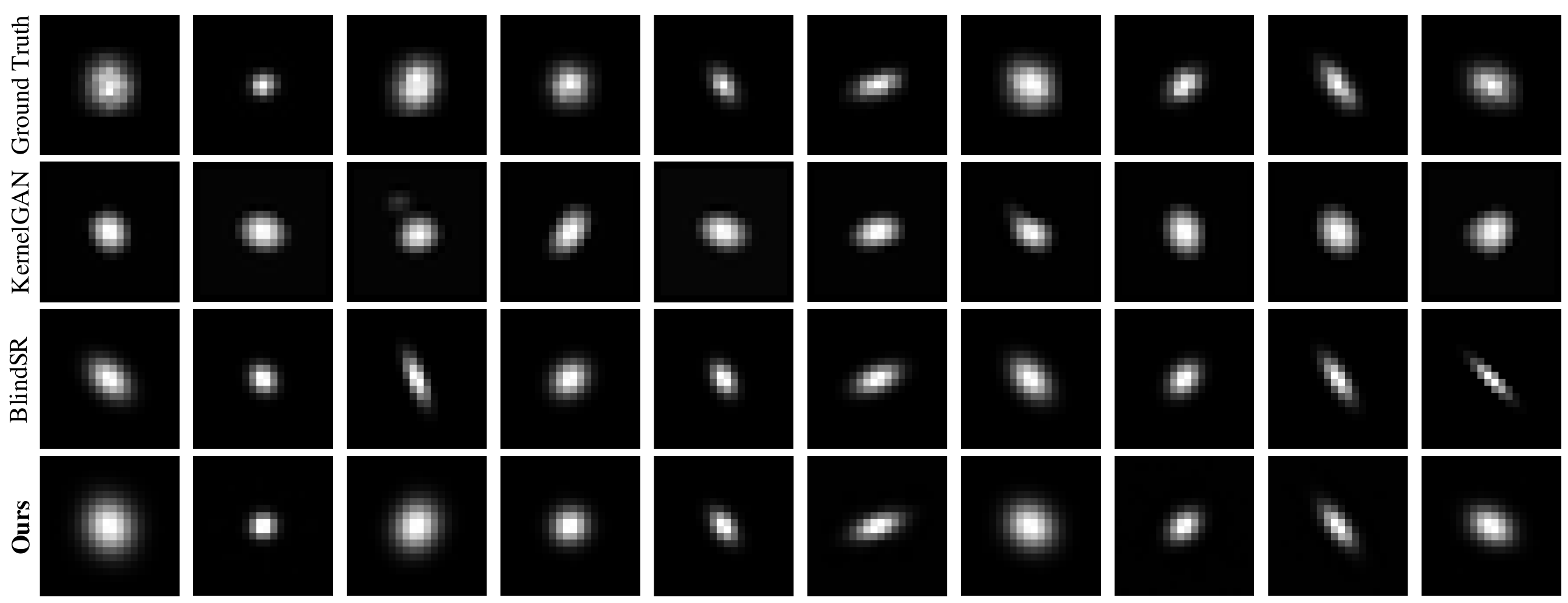}
\caption{Comparison of estimated degradation kernels on DIV2KRK\cite{bell2019kernelgan}}
\end{subfigure}
\caption{Visualizations of ground truth kernels and estimated degradation kernels by KernelGAN\cite{bell2019kernelgan}, BlindSR\cite{cornillere2019blind} and Ours on two datasets, (a) DIV2K-val and (b) DIV2KRK \cite{bell2019kernelgan}. The downsampling network of our KOALAnet is able to predict accurate degradation kernels on DIV2KRK as well as DIV2K-val.}
\label{fig:kernel_comparison}
\end{figure*}

To further analyze this aspect, we show the spatially-variant kernel estimations in an image mixed with in-focus and out-of-focus areas in Fig. \ref{fig:non_uniform_kernel_supp}. In \ref{fig:non_uniform_kernel_supp} (a), the estimated degradation kernel in the \textit{green box}, which is the in-focus area where the high frequency details must be restored, is highly similar to the ground truth degradation kernel. Comparing the \textit{red box} and \textit{blue box} in the smooth area, more accurate kernels are predicted in the region closer to the penguin (\textit{blue box}). In Fig. \ref{fig:non_uniform_kernel_supp} (b), we have visualized the cosine similarity map between the vectorized predicted per-pixel kernels and the ground truth blur kernel (\textit{red} denotes high similarity and \textit{blue} denotes low similarity). It can be clearly seen that the predicted degradation kernels are indeed \textit{spatially-variant} depending on the location in the image. Also, thanks to the large receptive field of our U-Net-based downsampling network, accurate blur kernels are predicted even in the smooth regions near the in-focus area (Fig. \ref{fig:non_uniform_kernel_supp} (b)). This helps to effectively handle the boundary areas between in-focus and completely out-of-focus areas so that these regions are not over-sharpened after SR. Note that conventional SR methods are unable to disentangle the degradation blur and the intended blur, and thus generates over-sharpened results in boundary regions (Fig. 1 in main paper).

\subsection{Comparison of Estimated Degradation Kernels}
\paragraph{Details on the kernel accuracy measurement.}
In Table 4 of the main paper, we compared the estimation accuracy of the degradation kernels generated by KernelGAN \cite{bell2019kernelgan}, BlindSR \cite{cornillere2019blind} and the downsampling network of our KOALAnet on the DIV2K-val testset with random anisotropic Gaussian degradations, and DIV2KRK \cite{bell2019kernelgan}. For each method, the degradation kernels were estimated from the input LR images degraded via the corresponding ground truth kernels. BlindSR\cite{cornillere2019blind} predicts three values, standard deviations $\sigma_{11}$ and $\sigma_{22}$ and the rotation angle $\theta$, of a bivariate Gaussian distribution to parametrize a Gaussian kernel. 
Since the peak of the estimated Gaussian kernel is located at the center pixel due to the size of the kernel ($15\times15$) being an odd number in the original implementation provided by the authors of BlindSR \cite{cornillere2019blind}, we calibrated the estimated kernels by convolving them with a $2\times2$ mean filter with 0.25 values. This yielded lower $l_2$ error. Furthermore, since the center of the ground truth kernels, $k_{gt}$, and the estimated kernels, $k$, can be different, we shifted $k$ and $k_{gt}$ in $x$ and $y$ directions to find the minimum $l_2$ error, as follows:
\begin{align} 
	{l_2} \ \text{error} =\min_{\Delta h, \Delta w}{\sum_{h,w} \lVert k_{gt}(h,w)-k(h\!+\!\Delta h, w\!+\!\Delta w) \rVert^{2}},
\label{eq:kernel_accuracy}
\end{align}
where $h,w$ represent the locations in x, y dimensions and $\Delta h, \Delta w$ is the shift along the x,y dimensions.

\medskip \noindent
\textbf{Visualization of the estimated kernels.}\quad
For the qualitative comparison of the estimated kernels, we visualized some examples of the ground truth kernels and the estimated kernels by KernelGAN \cite{bell2019kernelgan}, BlindSR \cite{cornillere2019blind} and the downsampling network of our KOALAnet on the DIV2K-val testset and DIV2KRK \cite{bell2019kernelgan} in Fig. \ref{fig:kernel_comparison}. As shown, our KOALAnet robustly estimates the latent degradation kernels from the LR images compared to other methods, even in DIV2KRK \cite{bell2019kernelgan} where noise is injected to the Gaussian kernels.

\section{Details of Complexity Evaluation}

\subsection{KOALAnet}
Our proposed framework is implemented on Python 3.6 with Tensorflow 1.13, and we used an NVIDIA Titan RTX for our experiments. The total number of filter parameters is 6.09M and 6.45M for the KOALAnet of $s=2$ and $s=4$, respectively. 

\subsection{Comparison to Existing Methods}
We provided a comparison on computational complexity in terms of the inference time and GFLOPs with other blind SR methods \cite{bell2019kernelgan,cornillere2019blind, gu2019ikc, shocher2018zssr} in Table 1 of the main paper. For all methods, GFLOPs is calculated only for the feed-forward paths. Furthermore, since ZSSR \cite{shocher2018zssr}, KernelGAN \cite{bell2019kernelgan} and BlindSR \cite{cornillere2019blind} are optimization-based methods, we take the number of iterations into consideration when computing the GFLOPs. The number of iterations needed for the optimization of ZSSR \cite{shocher2018zssr} can be different at each run even on the same test image (\textit{e}.\textit{g}., ``\textit{baby}'' in Set5), while the numbers of iterations of KernelGAN \cite{bell2019kernelgan} and BlindSR \cite{cornillere2019blind} are fixed. Thus for ZSSR, we obtained the average number of iterations from five repetitions each for scale factors 2 and 4, and then computed the GFLOPs using those numbers.

\begin{figure*}
\centering
\includegraphics[height=0.91\textheight]{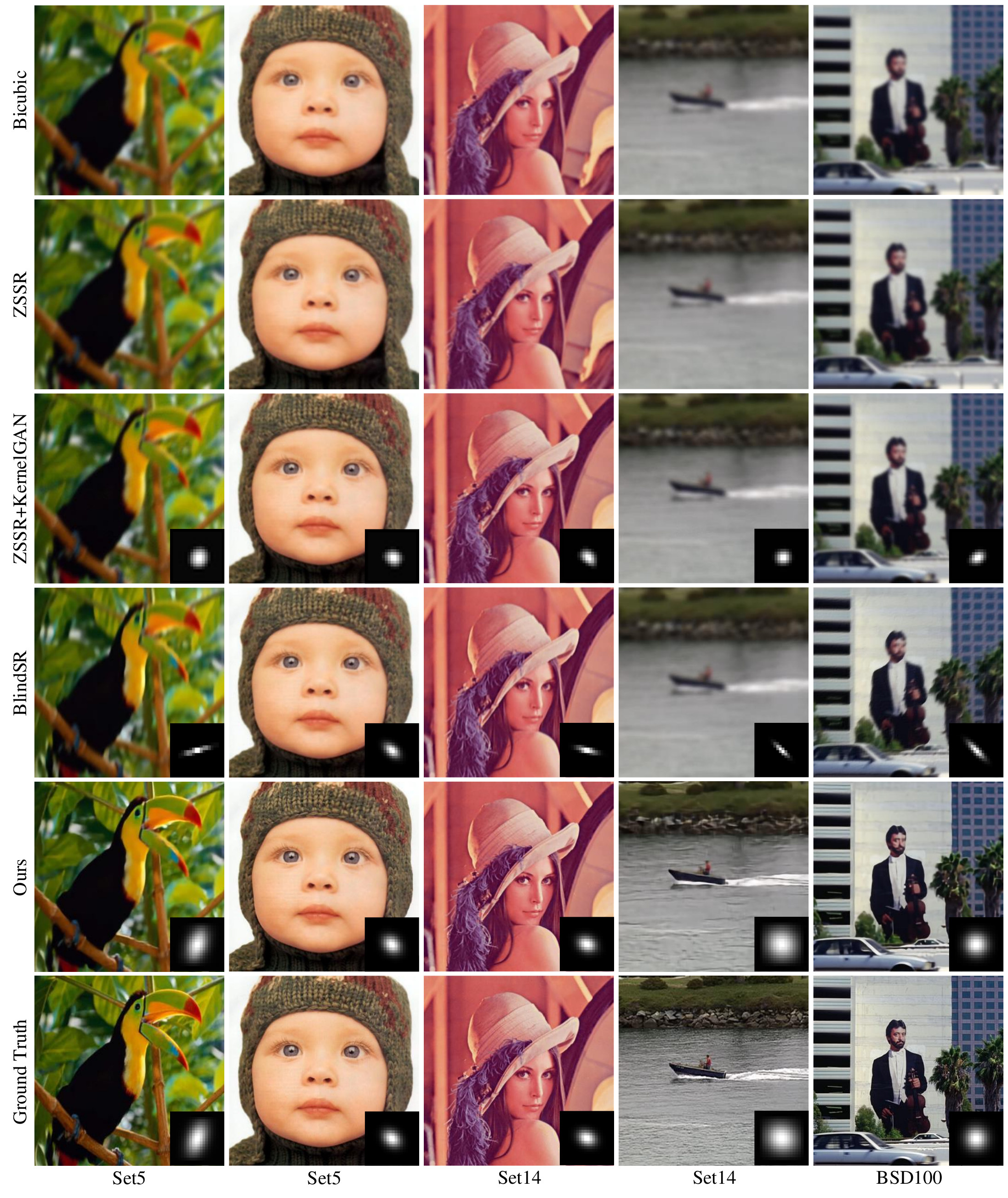}
\caption{Additional qualitative comparison with ZSSR \cite{shocher2018zssr}, ZSSR+KernelGAN \cite{bell2019kernelgan} and BlindSR \cite{cornillere2019blind} for scale factor 2 on Set5, Set14 and BSD100 datasets. The estimated (or ground truth) degradation kernels are placed on the bottom right corner for all applicable methods that estimate the degradation kernel. Our KOALAnet is able to predict accurate degradation kernels and generate sharp SR results on various datasets and degradation kernels.}
\label{fig:qual_comp_x2_1}
\end{figure*}

\begin{figure*}
\centering
\includegraphics[height=0.91\textheight]{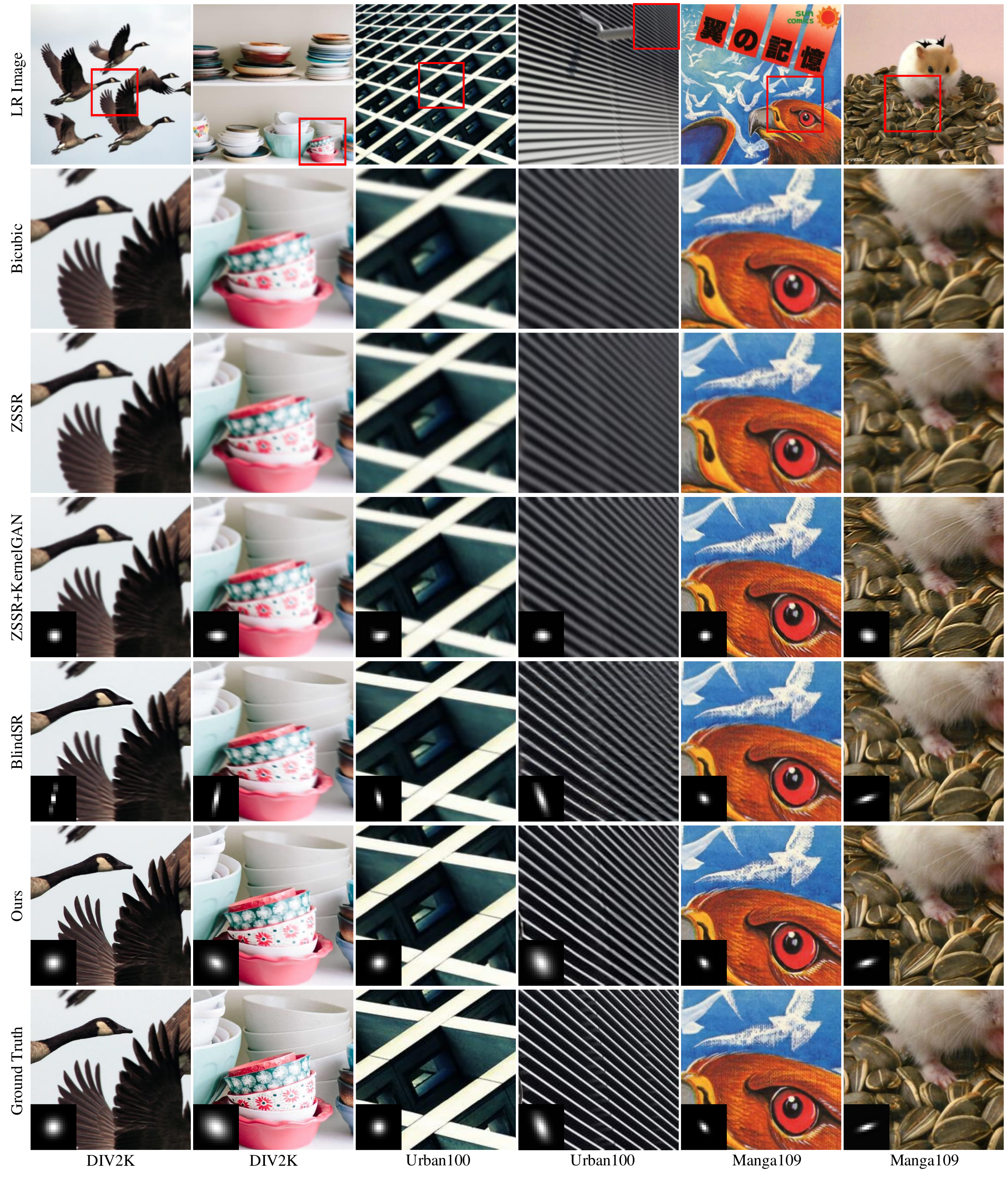}
\caption{Additional qualitative comparison with ZSSR \cite{shocher2018zssr}, ZSSR+KernelGAN \cite{bell2019kernelgan} and BlindSR \cite{cornillere2019blind} for scale factor 2 on DIV2K, Urban100 and Manga109 datasets. The estimated (or ground truth) degradation kernels are placed on the bottom left corner for all applicable methods that estimate kernel information. Our KOALAnet is able to predict accurate degradation kernels and generate sharp SR results on various datasets and degradation kernels.}
\label{fig:qual_comp_x2_2}
\end{figure*}

\begin{figure*}
\centering
\includegraphics[height=0.91\textheight]{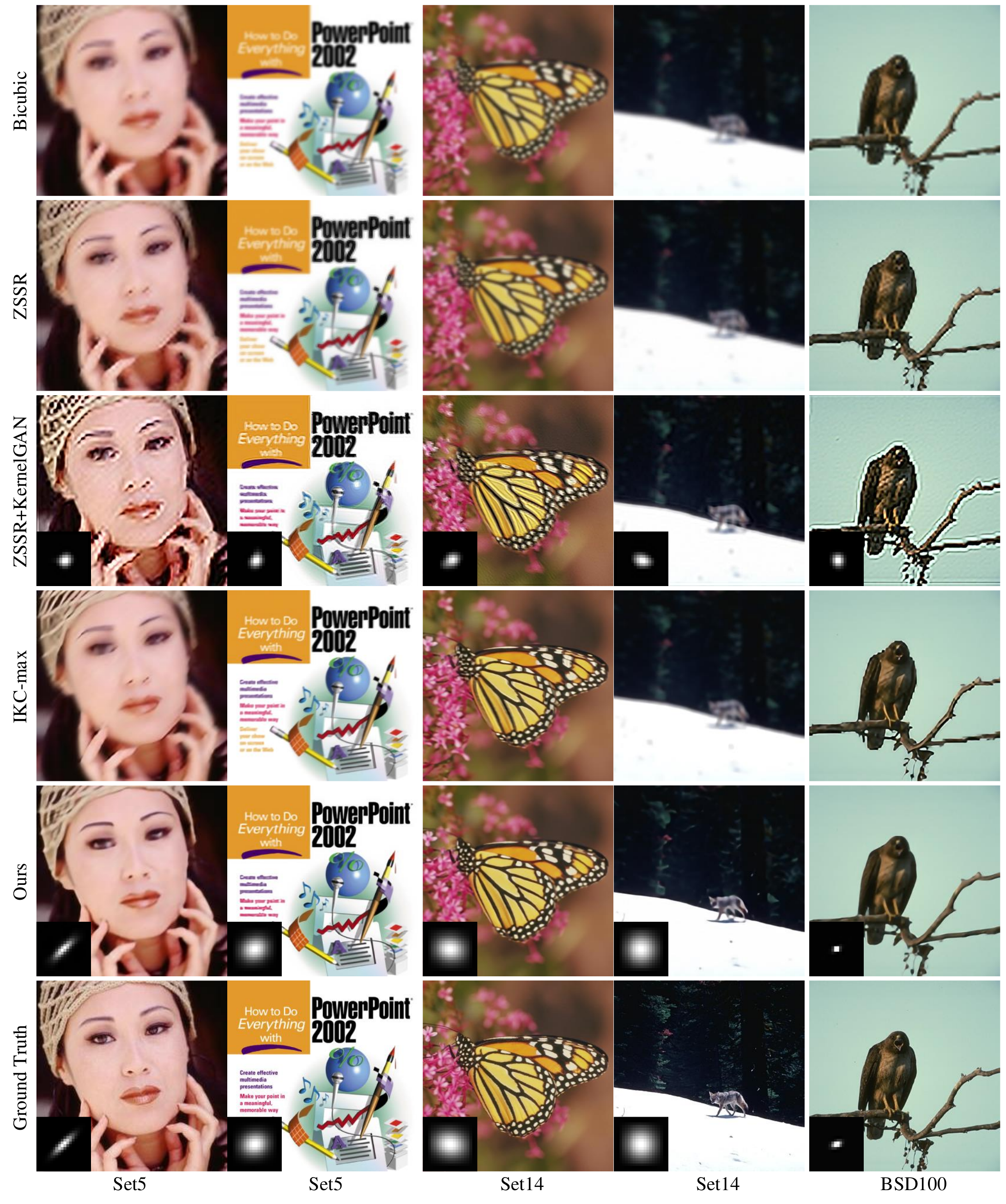}
\caption{Additional qualitative comparison with ZSSR \cite{shocher2018zssr}, ZSSR+KernelGAN \cite{bell2019kernelgan} and IKC \cite{gu2019ikc} for scale factor 4 on Set5, Set14 and BSD100 datasets. The estimated (or ground truth) degradation kernels are placed on the bottom left corner for all applicable methods that estimate the degradation kernel. We show the results yielding the best PSNR among seven iterations for IKC. Our KOALAnet is able to predict accurate degradation kernels and generate sharp SR results on various datasets and degradation kernels.}
\label{fig:qual_comp_x4_1}
\end{figure*}

\begin{figure*}
\centering
\includegraphics[height=0.91\textheight]{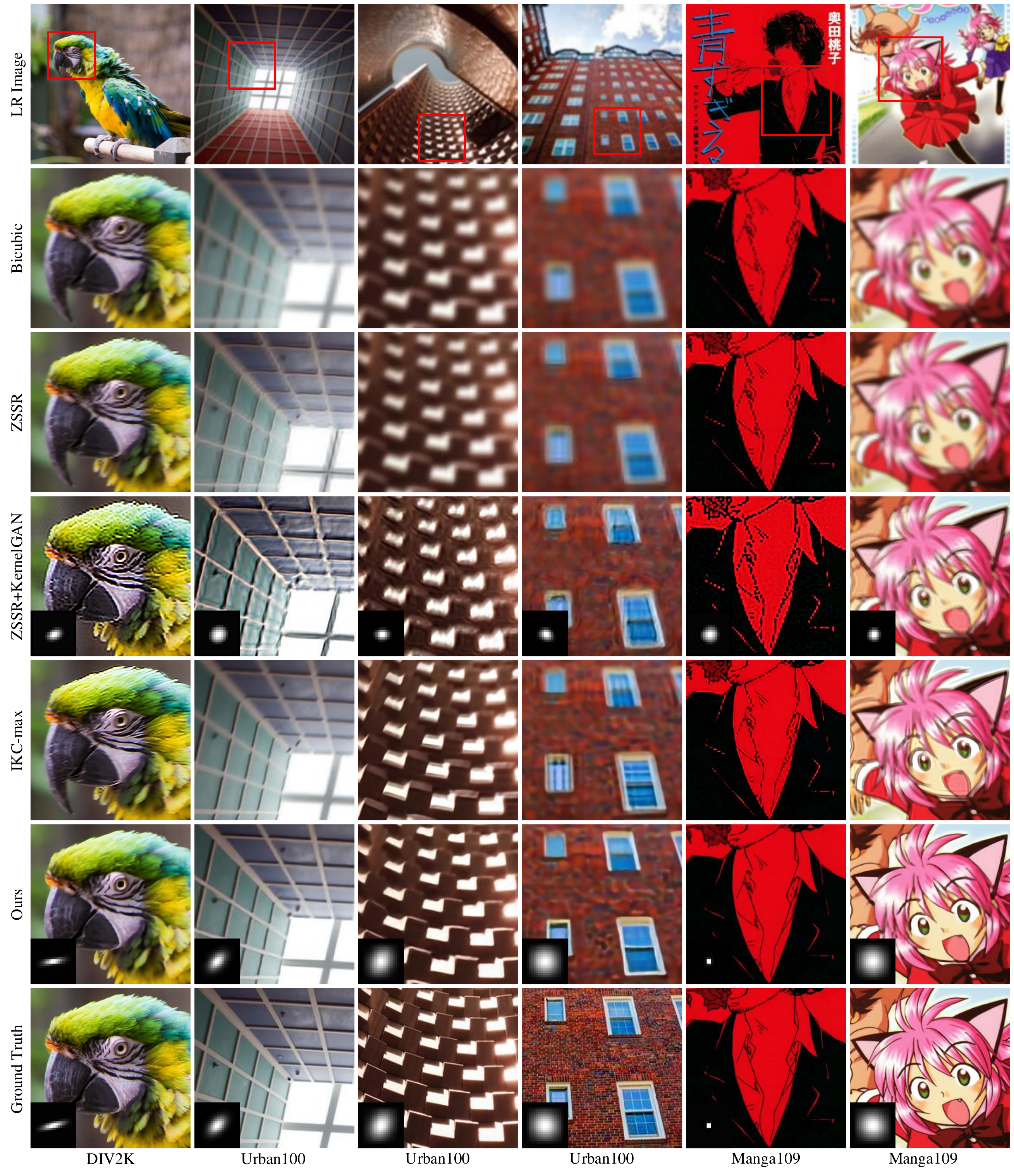}
\caption{Additional qualitative comparison with ZSSR \cite{shocher2018zssr}, ZSSR+KernelGAN \cite{bell2019kernelgan} and IKC \cite{gu2019ikc} for scale factor 4 on DIV2K, Urban100 and Manga109 datasets. The estimated (or ground truth) degradation kernels are placed on the bottom left corner for all applicable methods that estimate the degradation kernel. We show the results yielding the best PSNR among seven iterations for IKC. Our KOALAnet is able to predict accurate degradation kernels and generate sharp SR results on various datasets and degradation kernels.}
\label{fig:qual_comp_x4_2}
\end{figure*}

\section{Additional Qualitative Results}
We provide additional qualitative results on the random anisotropic Gaussian testsets in Fig. \ref{fig:qual_comp_x2_1} and \ref{fig:qual_comp_x2_2} for scale factor 2, and Fig. \ref{fig:qual_comp_x4_1} and Fig. \ref{fig:qual_comp_x4_2} for scale factor 4. The estimated degradation kernels are shown at the bottom right or the bottom left corner for all applicable methods that estimate the degradation kernel. Since IKC \cite{gu2019ikc} estimates the degradation kernels in a lower dimensionality (after PCA), they cannot be visualized along with the other degradation kernels estimated from BlindSR \cite{cornillere2019blind}, KernelGAN \cite{bell2019kernelgan} or Ours, from which actual degradation kernels can be generated. The ground truth degradation kernels are also shown at the bottom right or the bottom left corner of images or patches denoted as Ground Truth. In Fig. \ref{fig:qual_comp_x2_1} and Fig. \ref{fig:qual_comp_x4_1}, we show full images for comparisons on Set5, Set14 and BSD100, which have relatively low resolutions. In Fig. \ref{fig:qual_comp_x2_2} and Fig. \ref{fig:qual_comp_x4_2}, we crop the SR results to better visualize the difference for the readers for comparisons on DIV2K, Urban100 and Manga109, which have higher resolutions of near 2K. For IKC \cite{gu2019ikc}, we visualized the results yielding the best PSNR performance among seven iterations (IKC-max). As shown, our KOALAnet produces accurate SR results with sharper edges and realistic textures on various datasets and degradation kernels, even in examples with very high frequency regions. 

\end{appendices}

\end{document}